\documentclass[sigconf]{acmart}

\usepackage[utf8]{inputenc} 
\usepackage[T1]{fontenc}    
\usepackage{url}            
\usepackage{nicefrac}       

\usepackage{xcolor}
\usepackage{multirow}
\usepackage{tabularx}
\usepackage{adjustbox}
\usepackage{color, colortbl}
\definecolor{Gray}{gray}{0.9}

\usepackage{microtype}
\usepackage{graphicx}
\usepackage{subfigure}
\usepackage{amsfonts}
\usepackage{amsmath,bm}
\usepackage{booktabs}
\usepackage{mathtools}
\usepackage[most]{tcolorbox}

\usepackage{algorithmic}
\usepackage[ruled,vlined]{algorithm2e}
\usepackage{wrapfig}

\usepackage[font=small,skip=0pt]{caption}
\usepackage{listings}
\definecolor{codegreen}{rgb}{0,0.6,0}
\definecolor{codegray}{rgb}{0.5,0.5,0.5}
\definecolor{codepurple}{rgb}{0.58,0,0.82}
\definecolor{backcolour}{rgb}{0.95,0.95,0.92}
\lstdefinestyle{mystyle}{
    backgroundcolor=\color{backcolour},   
    commentstyle=\color{codegreen},
    keywordstyle=\color{magenta},
    numberstyle=\tiny\color{codegray},
    stringstyle=\color{codepurple},
    basicstyle=\ttfamily\footnotesize,
    breakatwhitespace=false,         
    breaklines=true,                 
    captionpos=b,                    
    keepspaces=true,                 
    numbers=left,                    
    numbersep=5pt,                  
    showspaces=false,                
    showstringspaces=false,
    showtabs=false,                  
    tabsize=2
}
\usepackage{pifont}

\usepackage{adjustbox}
\usepackage{color, colortbl}
\definecolor{Gray}{gray}{0.9}
\definecolor{babypink}{rgb}{0.96, 0.76, 0.76}
\definecolor{champagne}{rgb}{0.97, 0.91, 0.81}

\usepackage{listings}
\definecolor{codegreen}{rgb}{0,0.6,0}
\definecolor{codegray}{rgb}{0.5,0.5,0.5}
\definecolor{codepurple}{rgb}{0.58,0,0.82}
\definecolor{backcolour}{rgb}{0.95,0.95,0.92}
\lstdefinestyle{mystyle}{
    backgroundcolor=\color{backcolour},   
    commentstyle=\color{codegreen},
    keywordstyle=\color{magenta},
    numberstyle=\tiny\color{codegray},
    stringstyle=\color{codepurple},
    basicstyle=\ttfamily\footnotesize,
    breakatwhitespace=false,         
    breaklines=true,                 
    captionpos=b,                    
    keepspaces=true,                 
    numbers=left,                    
    numbersep=5pt,                  
    showspaces=false,                
    showstringspaces=false,
    showtabs=false,                  
    tabsize=2
}

\usepackage{hyperref}
\hypersetup{
    colorlinks=true,
    linkcolor=blue,
    citecolor=cyan,      
    urlcolor=cyan
    }

\newtheorem{example}{Example}
\newtheorem{lemma}{Lemma}
\newtheorem{remark}{Remark}

\newtheorem{theorem}{Theorem}
\newtheorem{assumption}{Assumption}
\newtheorem{definition}{Definition}

\AtBeginDocument{%
  }

\copyrightyear{2025}
\acmYear{2025}
\setcopyright{rightsretained}
\acmConference[KDD '25]{Proceedings of the 31st ACM SIGKDD Conference on
Knowledge Discovery and Data Mining V.1}{August 3--7, 2025}{Toronto, ON,
Canada}
\acmBooktitle{Proceedings of the 31st ACM SIGKDD Conference on KnowledgeDiscovery and Data Mining V.1 (KDD '25), August 3--7, 2025, Toronto, ON,
Canada}
\acmDOI{10.1145/3690624.3709337} 
\acmISBN{979-8-4007-1245-6/25/08}

\begin{document}

\title{Robust Fast Adaptation from Adversarially Explicit Task Distribution Generation}

\author{Qi (Cheems) Wang$^\dagger$}
\authornote{These authors contributed equally to this research.}
\affiliation{%
  \institution{Tsinghua University}
  \city{Beijing}
  \country{China}
}

\author{Yiqin Lv}
\authornotemark[1]
\affiliation{%
  \institution{Tsinghua University}
  \city{Beijing}
  \country{China}
  \email{xxx}}

\author{Yixiu Mao}
\authornotemark[1]
\affiliation{%
  \institution{Tsinghua University}
  \city{Beijing}
  \country{China}
  \email{xxx}}

\author{Yun Qu}
\affiliation{%
  \institution{Tsinghua University}
  \city{Beijing}
  \country{China}
  \email{xxx}}
  
\author{Yi Xu}
\affiliation{%
  \institution{Dalian University of Technology}
  \city{Dalian}
  \country{China}
  \email{xxx}}
  
\author{Xiangyang Ji}
\authornote{Correspondence: \texttt{cheemswang@mail.tsinghua.edu.cn}; \texttt{xyji@tsinghua.edu.cn}}
\affiliation{%
  \institution{Tsinghua University}
  \city{Beijing}
  \country{China}
  \email{xyji@tsinghua.edu.cn}}

\renewcommand{\shortauthors}{Wang et al.}

\begin{abstract}
Meta-learning is a practical learning paradigm to transfer skills across tasks from a few examples. 
Nevertheless, the existence of \textit{task distribution shifts} tends to weaken meta-learners' generalization capability, particularly when the training task distribution is naively hand-crafted or based on simple priors that fail to cover critical scenarios sufficiently.
Here, we consider explicitly generative modeling task distributions placed over task identifiers and propose robustifying fast adaptation from adversarial training.
Our approach, which can be interpreted as a model of a Stackelberg game, not only uncovers the task structure during problem-solving from an explicit generative model but also theoretically increases the adaptation robustness in worst cases. 
This work has practical implications, particularly in dealing with task distribution shifts in meta-learning, and contributes to theoretical insights in the field.
Our method demonstrates its robustness in the presence of task subpopulation shifts and improved performance over SOTA baselines in extensive experiments.
The code is available at the project site (\color{blue}{\url{https://sites.google.com/view/ar-metalearn}}).
\end{abstract}



\received[accepted]{16 November 2024}

\maketitle

\section{Introduction}

Deep learning has made remarkable progress in the past decade, ranging from academics to industry \citep{lecun2015deep}.
However, training deep learning models is generally time-consuming, and the previously trained model on one task might perform poorly in deployment when faced with unseen scenarios \citep{lesort2021understanding}.

Fortunately, meta-learning, or learning to learn, offers a scheme to generalize learned knowledge to unseen scenarios \citep{finn2017model,duan2016rl,hochreiter2001learning,hospedales2021meta}.
The strategy is to leverage past experience, extract meta knowledge as the prior, and utilize a few shot examples to transfer skills across tasks.
This way, we can avoid learning from scratch and quickly adapt the model to unseen but similar tasks, catering to practical demands, such as fast autonomous driving in diverse scenarios.
Due to these desirable properties, such a learning paradigm is playing an increasingly crucial role in building foundation models \citep{min2023recent,khan2022transformers,bai2023sequential,wang2023large}.
\begin{figure}
  \begin{center}
    \includegraphics[width=0.47\textwidth]{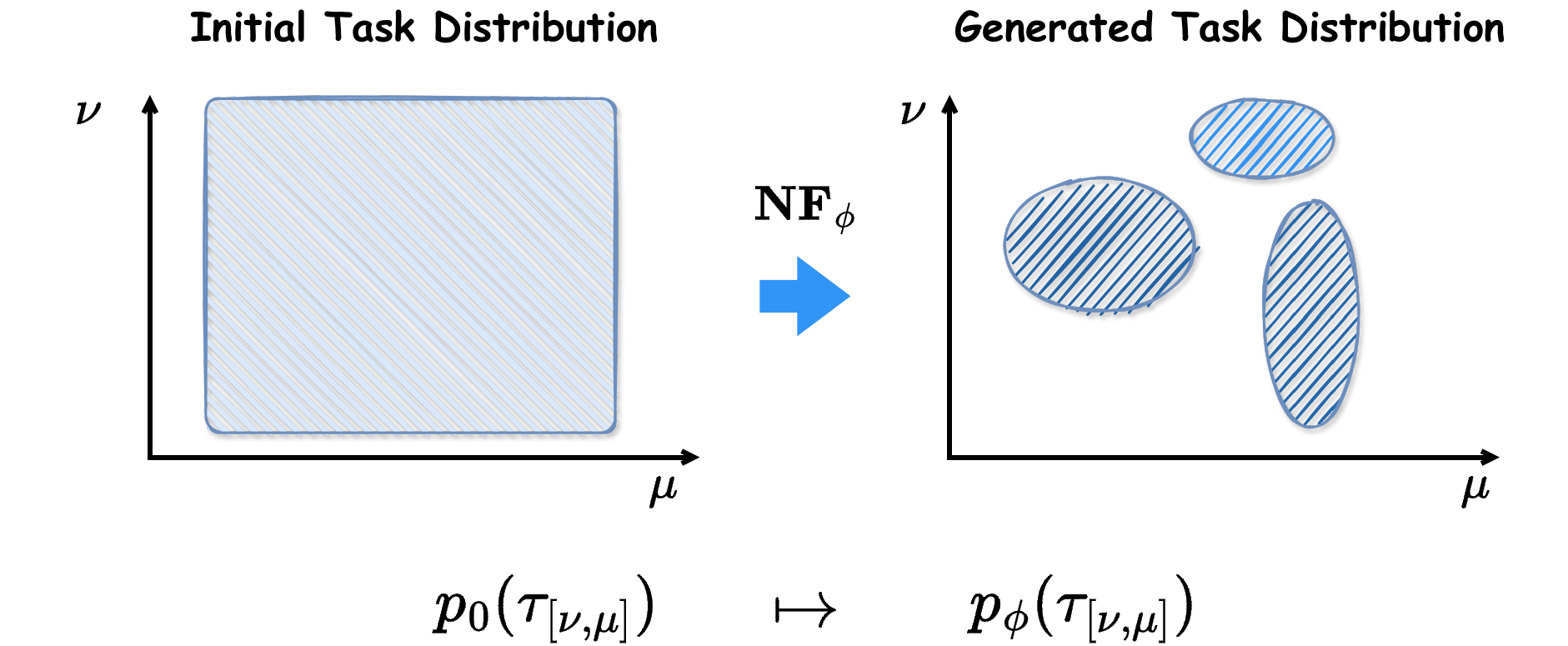}
  \caption{\textbf{Diagram of Generating Task Distribution as the Adversary in Meta-Learning.}
  Here, the initial task distribution $p_{0}(\tau)$ is a uniform distribution governed by two task identifiers $[\nu,\mu]$.
  Then, it is transformed into an explicit distribution $p_{\bm\phi}(\tau)$ with the help of normalizing flows $\texttt{\textbf{NF}}_{\bm\phi}$.}
  \vspace{-10pt}
  \label{adaptive_dist_demo}
  \end{center}
\end{figure}

\textbf{Literature Challenges:}
Despite the promising adaptation performance in meta-learning, several concerns remain.
Among them, the automatically task distribution design is under-explored and challenging in the field, which closely relates to the model's generalization evaluation \citep{yu2020meta,conklin2021meta}.

Overall, task identifiers configure the task, such as the topic type in the corpus for large language models \citep{brown2020language,wang2023large}, the amplitude and phase in sinusoid functions, or the degree of freedom in robotic manipulators \citep{faverjon1987local,anne2021meta}.
Most existing studies adopt simple prior, such as uniform distributions over task identifiers \citep{finn2017model,garnelo2018conditional,rusu2018meta}, or hand-crafted distributions, which heavily rely on domain-specific knowledge difficult to acquire.

Some scenarios even pose more realistic demands for task distributions.
In testing an autopilot system, an ideal task distribution deserves more attention on traffic accidents or even generates some while covering typical cases \citep{tan2021scenegen,rempe2022generating}.
Similar circumstances also occur during domain randomization for embodied robots \citep{mehta2020active}.
These imply that the shift between commonly used task distributions, such as uniform, and the expected testing distributions raises robustness issues and probably causes catastrophic failures when adapting to risk-sensitive scenarios \citep{mao2023supported}.

\textbf{Proposed Solutions:}
Rather than exploring fast adaptation strategies, we turn to \textit{explicitly create task distribution shifts at a certain level and characterize robust fast adaptation with a Stackelberg game} \citep{osborne2004introduction}.
To this end, we utilize normalizing flows to parameterize the distribution adversary in Figure \ref{adaptive_dist_demo} for task distribution generation and the meta learner for fast adaptation in the presence of distribution shifts.

Importantly, we constitute the solution concept, adopt the alternative gradient descent ascent to approximately compute the equilibrium \citep{kreps1989nash}, and conduct theoretical analysis.
The optimization process can be translated as \textit{fast adaptation robustification through adversarially explicit task distribution generation}.

\textbf{Outline \& Primary Contributions:}
The remainder starts with related work in Section \ref{liter}.
We define the notation and recap fundamentals in Section \ref{prelim_sec}.
Then, we present the game-theoretical framework to handle constrained task distribution shifts and robustify fast adaptation in Section \ref{method_sec}.
The quantitative analysis is conducted in Section \ref{exp_sec}, followed by conclusions and limitations.
In primary, our contributions are:
\begin{itemize}
    \item This work translates the robust fast adaptation under distribution shifts into a Stackelberg game \citep{von1952theory}.
    To reveal task structures during problem-solving, we explicitly generate the task distribution with normalizing flows over task identifiers and optimize the meta-learner in an adversarial way.
    \item In theoretical analysis and tractable optimization, we constitute the solution concept \textit{w.r.t.} fast adaptation, approximately solve the game using alternating stochastic gradient descent, and perform convergence and generalization analysis under certain conditions.
\end{itemize}
Extensive experimental results show that our approach can reveal adaptation-related structures in the task space and achieve robustness improvement in task subpopulation shifts.

\section{Literature Review}\label{liter}
The past few years have developed a large body of work on skill transfer across tasks or domain generalization in different ways \citep{hospedales2021meta,xiao2023energy,xiao2022learning,xiao2024any}.
This section overviews the field regarding meta-learning and adaptation robustness.

\textbf{Meta Learning.}
Meta learning is a learning paradigm that considers a distribution over tasks.
The key is to pursue strategies for leveraging past experiences and distilling extracted knowledge into unseen tasks with a few shots of examples \citep{munkhdalai2017meta,hospedales2021meta,chimeta}.
Currently, there are various families of meta-learning methods.
The optimization-based ones, like model agnostic meta-learning (MAML)  \citep{finn2017model} and its extensions \citep{grant2018recasting,rajeswaran2019meta,fallah2020convergence,wang2020global}, aim at finding a good meta-initialization of model parameters for adapting to all tasks via gradient descent.
The deep metrics methods optimize the task representation in a metric space and are superior in few-shot image classification tasks \citep{snell2017prototypical,li2019revisiting,allen2019infinite,huang2021local,yang2021mining}.
Typical context-based methods, e.g., neural processes (NPs) and variants \citep{garnelo2018conditional,wang2020doubly,garnelo2018neural,wang2022bridge,wang2022model,kim2018attentive,gordon2019convolutional,wang2022learning,requeima2019fast,shen2023episodic}, constitute the deep latent variable model as the stochastic process to accomplish tasks.
Besides, memory-augmented networks \citep{santoro2016meta}, hyper-networks \citep{ha2016hypernetworks}, and so forth are designed for meta-learning purposes.

\textbf{Robustness in Meta Learning.}
In most previous work, the task distribution is fixed in the training set-up.
In order to robustify the fast adaptation performance, a couple of learning strategies or principles emerge.
Increasing the robustness to worst cases is a commonly seen consideration in adaptation, and these scenarios include input noise, parameter perturbation, and task distributions \citep{olds2015global,kurakin2016adversarial,liu2018security,zhang2020cautious,chi2021tohan,tay2022efficient,li2022defensive}.
To alleviate the effects of adversarial examples in few-shot image classification, \citet{goldblum2019robust} meta-train the model in an adversarial way.
To handle the distribution mismatch between training and testing tasks, \citet{zhang2021adaptive} adopt the adaptive risk minimization principle to enable fast adaptation.
\citet{wang2023sim} propose to optimize the expected tail risk in meta-learning and witness the increase of robustness in proportional worst cases.
Ours is a variant of a distributionally robust framework \citep{wiesemann2014distributionally}, and we seek equilibrium for fast adaptation.

\textbf{Task Distribution Studies in Meta Learning.}
Task distributions are directly related to the generalization capability of meta-learning models, attracting increasing attention recently.
Aiming to alleviate task overfitting, \citet{rajendran2020meta,yao2021improving,murty2021dreca,ni2021data} enrich the task space with augmentation techniques.
Task relatedness can improve generalization across tasks, \citet{fifty2021efficiently} devise an efficient strategy to group tasks in multi-task training.
In \citep{liu2020adaptive,yao2021meta}, neural task samplers are developed to schedule the probability of task sampling in the context of few-shot classification.
To increase the fidelity of generated tasks, \citet{wu2022adversarial} adopt the task representation model and constructs the up-sampling network for meta-training task augmentation.
To reduce the required tasks, \citep{yao2021meta,lee2022set} take the task interpolation strategy and shows that the interpolation strategy outperforms the standard set-up.
Distinguished from the above, this work takes more interest in explicitly understanding task identifier structures concerning learning performance and cares about fast adaptation robustness under subpopulation shift constraints.
Optimizing the task distribution might reserve the potential to improve generative performance in large models \citep{chiunveiling}.

\section{Preliminaries}\label{prelim_sec}

\textbf{Notation.}
Throughout this paper, we use $p(\tau)$ to denote the task distribution with $\mathcal{T}$ the task domain.
Here, $\mathcal{D}_{\tau}$ represents the meta dataset with a sampled task $\tau$.
With the model parameter domain $\bm\Theta$ and the support/query dataset construction, e.g., $\mathcal{D}_{\tau}=\mathcal{D}_{\tau}^{S}\cup \mathcal{D}_{\tau}^{Q}$, the risk function in meta-learning is a real-value function
$\mathcal{L}:\mathcal{T}\times\bm\Theta\mapsto\mathbb{R}$.

As an example, $\mathcal{D}_{\tau}$ consists of data points $\{(x_i,y_i)\}_{i=1}^{m+n}$ in few shot regression, and it is mostly split into the support dataset $\mathcal{D}_{\tau}^{S}$ for fast adaptation and query dataset $\mathcal{D}_{\tau}^{Q}$ for evaluation.

\subsection{Problem Statement}

To begin with, we revisit a couple of commonly-used risk minimization principles for meta-learning as follows.

\textbf{Standard Meta-Learning Optimization Objective.}
We consider the meta-learning problem within the expected risk minimization principle in the statistical learning theory \citep{vapnik1999nature}. 
This results in the objective as Eq. (\ref{meta_obj}), and we execute optimization in the form of task batches in implementation.
\begin{equation}\label{meta_obj}
        \min_{\bm\theta\in\bm\Theta}\mathbb{E}_{p(\tau)}\Big[\mathcal{L}(\mathcal{D}_{\tau}^{Q},\mathcal{D}_{\tau}^{S};\bm\theta)\Big]
\end{equation}
Here, $\bm\theta$ refers to the meta-learning model parameters for meta knowledge and fast adaptation.
The risk function depends on specific meta-learning methods.
For example, in MAML, the form can be $\mathcal{L}(\mathcal{D}_{\tau}^{Q},\mathcal{D}_{\tau}^{S};\bm\theta):=\mathcal{L}(\mathcal{D}_{\tau}^{Q};\bm\theta-\lambda\nabla_{\bm\theta}\mathcal{L}(\mathcal{D}_{\tau}^{S};\bm\theta))$ in regression, where the gradient update with the learning rate $\lambda$ in the bracket reflects fast adaptation.

\textbf{Distributionally Robust Meta Learning Optimization Objective.}
Recently, tail risk minimization has been adopted for meta-learning, effectively alleviating the effects towards fast adaptation in task distribution shifts \citep{wang2023sim}.
In detail, we can express the optimization objective as Eq. (\ref{dist_robust_meta_learning_1}) in the presence of the constrained distribution $p_{\alpha}(\tau;\bm\theta)$, which characterizes the $(1-\alpha)$ proportional $\bm\theta$-dependent worst cases in the task space.
\begin{equation}
\min_{\bm\theta\in\bm\Theta}\mathbb{E}_{p_{\alpha}(\tau;\bm\theta)}
        \Big[\mathcal{L}(\mathcal{D}_{\tau}^{Q},\mathcal{D}_{\tau}^{S};\bm\theta)
        \Big]
\label{dist_robust_meta_learning_1}
\end{equation}

It is worth noting that $p_{\alpha}(\tau;\bm\theta)$ is non-differentiable and $\theta$-dependent with no closed-form. 
Meanwhile, the worst-case optimization for meta-learning in Eq. (\ref{maxmin_meta_obj}) can be treated as a particular instance of Eq. (\ref{dist_robust_meta_learning_1}) when $\alpha$ sufficiently approaches $1$.
\begin{equation}\label{maxmin_meta_obj}
        \min_{\bm\theta\in\bm\Theta}\max_{\tau\in\mathcal{T}}\mathcal{L}(\mathcal{D}_{\tau}^{Q},\mathcal{D}_{\tau}^{S};\bm\theta)
\end{equation}

Through tail risk minimization, the model's adaptation robustness can be enhanced \textit{w.r.t.} the proportional worst scenarios \citep{collins2020task,wang2023sim}.

\subsection{Two-Player Stackelberg Game}
Before detailing our approach, it is necessary to describe elements in a two-player, non-cooperative Stackelberg game \citep{von1952theory}. 

Let us assume two competitive players are involved in the game 
$\Gamma:=\langle\{\mathcal{P}_1,\mathcal{P}_2\},\{\bm\theta\in\bm\Theta,\bm\phi\in\bm\Phi\},\mathcal{J}(\bm\theta,\bm\phi)\rangle,$
where the meta learner as the leader $\mathcal{P}_1$ makes a decision first in the domain $\bm\Theta$ while the distribution adversary as the follower $\mathcal{P}_2$ tries to deteriorate the leader decision's utility in the domain $\bm\Phi$.
We refer to $\mathcal{J}(\bm\theta,\bm\phi)$ as the continuous risk function of the leader $\mathcal{P}_1$, and that of the follower $\mathcal{P}_2$ corresponds to the negative form $-\mathcal{J}(\bm\theta,\bm\phi)$.
Without loss of generality, all the players are rational and try to minimize risk functions in the game.

\section{Task Robust Meta Learning under Distribution Shift Constraints}\label{method_sec}
\begin{figure*}[t]
\begin{center}
\centerline{\includegraphics[width=0.9\textwidth]{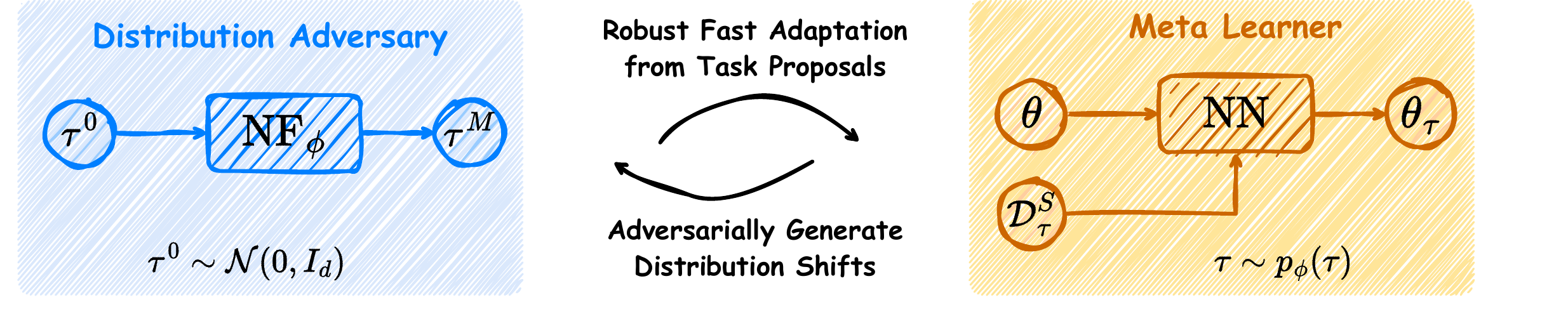}}
\caption{\textbf{Diagram of Adversarially Task Robust Meta Learning.}
The proposed framework consists of two players, the distribution adversary and the meta player, in the game of meta-learning.
\textbf{On the left side of the figure:} the distribution adversary seeks to transform the distribution from an initial task distribution, e.g., $\mathcal{N}(0,I_d)$ or $\mathcal{U}[a,b]$, via the neural network parameterized by $\bm\phi$ with the purpose of deteriorating meta player's fast adaptation performance.
\textbf{On the right side of the figure:} the meta player parameterized by $\bm\theta$ attempts to learn robust strategies for fast adaptation in sampled worst-case tasks (MAML algorithm \citep{finn2017model} as an illustration).
}
\label{adapt_dist_rob_meta}
\end{center}
\end{figure*}

This section starts with the game-theoretic framework for meta-learning, followed by approximate optimization.
Figure \ref{adapt_dist_rob_meta} shows a diagram of the constructed Stackelberg game.
Then we perform theoretical analysis \textit{w.r.t.} our approach.

\subsection{Generate Task Distribution within A Game-Theoretic Framework}

As part of an indispensable element in meta-learning, the task distribution is mostly set to be uniform or manually designed from the heuristics.
Such a setup hardly identifies a subpopulation of tasks that are tough to resolve in practice and fails to handle task distribution shifts.

In contrast, this paper considers an explicit task distribution to capture along with the learning progress and then automatically creates task distribution shifts for the meta-learner to adapt robustly.
Our framework can be categorized as curriculum learning \citep{bengio2009curriculum}, but there places a constraint over the distribution shift in optimization.

\textbf{Adversarially Task Robust Optimization with Distribution Shift Constraints.}
Now, we translate the meta-learning problem, namely generative task distributions for robust adaptation, into a min-max optimization problem:
\begin{equation}\label{game_meta_obj}
        \begin{split}
        \min_{\bm\theta\in\bm\Theta}\max_{\bm\phi\in\bm\Phi}\mathcal{J}(\bm\theta,\bm\phi):=\mathbb{E}_{p_{\bm\phi}(\tau)}\Big[\mathcal{L}(\mathcal{D}_{\tau}^{Q},\mathcal{D}_{\tau}^{S};\bm\theta)\Big],
        \\
        \textit{s.t.}
        \
        D_{KL}\Big[p_{0}(\tau)\parallel p_{\bm\phi}(\tau)\Big]\leq\delta,
        \end{split}
\end{equation}
where the constraint term defines the maximum distribution shift to tolerate in meta training.

Equivalently, we can rewrite the above optimization objective in the form of unconstrained one with the help of a Lagrange multiplier $\lambda\in\mathbb{R}^{+}$:
\begin{equation}\label{game_meta_obj}
        \begin{split}
        \min_{\bm\theta\in\bm\Theta}\max_{\bm\phi\in\bm\Phi}\mathcal{J}(\bm\theta,\bm\phi):=\mathbb{E}_{p_{\bm\phi}(\tau)}\Big[\mathcal{L}(\mathcal{D}_{\tau}^{Q},\mathcal{D}_{\tau}^{S};\bm\theta)\Big]
        \\
        -\lambda\Big[D_{KL}\Big[p_{0}(\tau)\parallel p_{\bm\phi}(\tau)\Big]-\delta\Big].
        \end{split}
\end{equation}
The above can be further simplified as:
\begin{equation}\label{main_obj}
    \begin{split}
        \min_{\bm\theta\in\bm\Theta}\max_{\bm\phi\in\bm\Phi}\mathcal{J}(\bm\theta,\bm\phi):=\mathbb{E}_{p_{\bm\phi}(\tau)}\Big[\mathcal{L}(\mathcal{D}_{\tau}^{Q},\mathcal{D}_{\tau}^{S};\bm\theta)\Big]
        +\lambda\mathbb{E}_{p_{0}(\tau)}\Big[\ln p_{\bm\phi}(\tau)\Big],
    \end{split}
\end{equation}
where the constant terms, e.g., $\lambda\delta\in\mathbb{R}^{+}$ and $\mathbb{E}_{p_{0}(\tau)}\Big[\ln p_{0}(\tau)\Big]$ are eliminated.

As previously mentioned, the role of the distribution adversary attempts to transform the initial task distribution into one that raises challenging task proposals with higher probability.
Such a setup drives \textit{the evolution of task distributions via adaptively shifting task sampling chance under constraints}, which can be more crucial for generalization across risky scenarios.
The term $D_{KL}\Big[p_{0}(\tau)\parallel p_{\bm\phi}(\tau)\Big]$ inside Eq. (\ref{game_meta_obj}) works as regularization to avoid the mode collapse in the generative task distribution.
In Figure \ref{adapt_dist_rob_meta}, the goal of the meta learner retains that of traditional meta-learning, while the distribution adversary continually generates the task distribution shifts along optimization processes.
\begin{assumption}[Lipschitz Smoothness and Compactness]\label{assum_lipschitz}
The adversarially task robust meta-learning optimization objective $\mathcal{J}(\bm\theta,\bm\phi)$ is assumed to satisfy
\begin{enumerate}
    \item 
    $\mathcal{J}(\bm\theta,\bm\phi)$ with $\forall[\bm\theta,\bm\phi]\in\bm\Theta\times\bm\Phi$ belongs to the class of twice differentiable functions $\mathbb{C}^2$.
    \item 
    The norm of block terms inside Hesssian matrices $\nabla^2\mathcal{J}(\bm\theta,\bm\phi)$ is bounded, meaning that $\forall[\bm\theta,\bm\phi]\in\bm\Theta\times\bm\Phi$:
    $$\sup\{||\nabla^2_{\bm\theta,\bm\theta}\mathcal{J}||,||\nabla^2_{\bm\theta,\bm\phi}\mathcal{J}||,||\nabla^2_{\bm\phi,\bm\phi}\mathcal{J}||\}\leq L_{\text{max}}.$$
    \item 
    The parameter spaces $\bm\Theta\subseteq\mathbb{R}^{d_1}$ and $\bm\Phi\subseteq\mathbb{R}^{d_2}$ are compact with $d_1$ and $d_2$ respectively dimensions of model parameters for two players.
\end{enumerate}
\end{assumption}

\begin{example}[Adversarially Task Robust MAML, AR-MAML]\label{example_armaml}
Given the parameterized task distribution $p_{\bm\phi}(\tau)$, the risk function $\mathcal{L}$ and the learning rate $\gamma$ in the inner loop of MAML \citep{finn2017model}, the adversarially task robust MAML corresponds to the following optimization problem:
\begin{equation}
    \begin{split}
        \min_{\bm\theta\in\bm\Theta}\max_{\bm\phi\in\bm\Phi}\mathbb{E}_{p_{\bm\phi}(\tau)}\Big[\mathcal{L}\left(\mathcal{D}_{\tau}^{Q};\bm\theta-\gamma
        \nabla_{\bm\theta}\mathcal{L}(\mathcal{D}_{\tau}^{S};\bm\theta)
        \right)\Big]+\lambda\mathbb{E}_{p_{0}(\tau)}\Big[\ln p_{\bm\phi}(\tau)\Big]
    \end{split}
    \label{AR_MAML_obj}
\end{equation}
where $\mathcal{D}_{\tau}^{S}$ is used for the inner loop with $\mathcal{D}_{\tau}^{Q}$ used for the outer loop.
\end{example}

\begin{example}[Adversarially Task Robust CNP, AR-CNP]\label{example_arcnp}
    Given the parameterized task distribution $p_{\bm\phi}(\tau)$, the risk function $\mathcal{L}$, and the conditional neural process \citep{garnelo2018conditional}, the adversarially task robust CNP can be formulated as follows:
    \begin{equation}
        \begin{split}
            \min_{\bm\theta\in\bm\Theta}\max_{\bm\phi\in\bm\Phi}\mathbb{E}_{p_{\bm\phi}(\tau)}\Big[\mathcal{L}(\mathcal{D}_{\tau}^{Q};z,\bm\theta_2)\Big]+\lambda\mathbb{E}_{p_{0}(\tau)}\Big[\ln p_{\bm\phi}(\tau)\Big],
        \\
        \text{s.t.}
        \
        z=h_{\bm\theta_1}(\mathcal{D}_{\tau}^{S})
        \
        \text{with}
        \
        \bm\theta=\{\bm\theta_1,\bm\theta_2\},
        \end{split}
    \end{equation}
where $\bm\theta_1$ and $\bm\theta_2$ are respectively a set encoder and the decoder networks.
\end{example}

Here, we take two typical methods, e.g., MAML \citep{finn2017model} and CNP \citep{garnelo2018conditional}, to illustrate the meta learner within the adversarially task robust framework, see Examples \ref{example_armaml}/\ref{example_arcnp} for details.

\textbf{Explicit Task Distribution Adversary Construction with Normalizing Flows.}
Learning to transform the task distribution is treated as a generative process: $\bm\Phi:\mathcal{T}\to\mathcal{T}\subseteq\mathbb{R}^d$ in this paper.
Admittedly, there already exist a collection of generative models to achieve the goal of generating task distributions, e.g., variational autoencoders \citep{kingma2013auto,rezende2014stochastic}, generative adversarial networks \citep{goodfellow2020generative}, and normalizing flows \citep{rezende2014stochastic}.

Among them, we propose to utilize the normalizing flow \citep{rezende2014stochastic} to achieve \textit{due to its tractability of the exact log-likelihood, flexibility in capturing complicated distributions, and a direct understanding of task structures}.
The basic idea of normalizing flows is to transform a simple distribution into a more flexible distribution with a series of invertible mappings $\mathcal{G}=\{g_i\}_{i=1}^{M}$, where $g_i:\mathcal{T}\to\mathcal{T}\subseteq\mathbb{R}^d$ indicates the smooth invertible mapping.
We refer to these mappings implemented in the neural networks as $\texttt{NN}_{\bm\phi}$ afterward.
Specifically, with the base distribution $p_0(\tau)$ and a task sample $\tau^0$, the model applies the above mappings to $\tau^0$ to obtain $\tau^M$.
\begin{equation}
    \begin{split}
        \tau^M=g_{M}\circ\dots g_{2}\circ g_{1}(\tau^0)=\texttt{NN}_{\bm\phi}(\tau^0)
    \end{split}
    \label{normalizing_flows_eq}
\end{equation}
In this way, the task distribution of interest is adaptive and adversarially exploits information from the shifted task distributions.
The density function after transformations can be easily computed with the help of functions' Jacobians:
\begin{equation}
    \begin{split}
        \ln p_{\bm\phi}(\tau^{M})=\ln p_{0}(\tau^{0})
        -\sum_{i=1}^{M}\ln\left|\det\frac{\partial g_{i}}{\partial \tau^{i-1}}\right|.
    \end{split}
    \label{nf_logdeterm}
\end{equation}

\begin{definition}[$(\ell_1, \ell_2)$-bi-Lipschitz Function]\label{append:def_bi_lip}
    An invertible function $g:x\subseteq\mathcal{X}\mapsto x\subseteq\mathcal{X}$, is said to be $(\ell_1, \ell_2)$-bi-Lipschitz if $\forall \{x_1,x_2\}\in\mathcal{X}$, the following conditions hold:
    $$|g(x_1)-g(x_2)|\leq\ell_{2}|x_1-x_2|\quad\text{and}\quad|g^{-1}(x_1)-g^{-1}(x_2)|\leq\ell_{1}|x_1-x_2|.$$
\end{definition}

As the normalizing flow function is invertible, the \textbf{Definition} \ref{append:def_bi_lip} is to describe the Lipschitz continuity in bi-directions.

\subsection{Solution Concept \& Explanations}
This work separates players regarding the decision-making order, and the optimization procedure is no longer a simultaneous game.
The nature of Stackelberg game enables us to technically express the studied asymmetric bi-level optimization problem as:
\begin{equation}
    \begin{split}
        \min_{\bm\theta\in\bm\Theta}\mathcal{J}(\bm\theta,\bm\phi),
        \
        \textit{s.t.}
        \
        \bm\phi\in\mathcal{S}(\bm\theta)
    \end{split}
\end{equation}
with the $\bm\theta$-dependent conditional subset $\mathcal{S}(\bm\theta):=\{\bm\phi\in\bm\Phi\vert\mathcal{J}(\bm\theta,\bm\phi)\geq\max_{\bm\phi\in\bm\Phi}\mathcal{J}(\bm\theta,\bm\phi)\}$.
This suggests the variables $\bm\theta$ and $\bm\phi$ are entangled in optimization.

Moreover, we can define the resulting equilibrium as a local minimax point \citep{jin2020local} in adversarially task robust meta-learning, due to the non-convex optimization practice. 
\begin{definition}[Local Minimax Point]\label{local_se}
    The solution $\{\bm\theta_*,\bm\phi_*\}$ is called local Stackelberg equilibrium when satisfying two conditions:
    (1) $\bm\phi_*\in\bm\Phi^{\prime}\subset\bm\Phi$ is the maximum of the function $\mathcal{J}(\bm\theta_*,\cdot)$ with $\bm\Phi^{\prime}$ a neighborhood;
    (2) $\bm\theta_*\in\bm\Theta^{\prime}\subset\bm\Theta$ is the minimum of the function $\mathcal{J}(\bm\theta,g(\bm\theta))$ with $g(\bm\theta)$ the implicit function of $\nabla_{\bm\phi}\mathcal{J}(\bm\theta,\bm\phi)=0$ in the neighborhood $\bm\Theta^{\prime}$.
\end{definition}

Moreover, there exists a clearer interpretation \textit{w.r.t.} the sequential optimization process and the equilibrium in the \textbf{Definition} \ref{local_se}.
The meta learner as the leader first optimizes its parameter $\bm\theta$.
Then the distribution adversary as the follower updates the parameter $\bm\phi$ and explicitly generates the task distribution proposal to challenge adaptation performance.
In other words, we expect that meta learners can benefit from generative task distribution shifts regarding the adaptation robustness.

\begin{remark}[Entropy of the Generated Task Distribution] \label{remark:entropy}
    Given the generative task distribution $p_{\bm\phi_*}(\tau)$, we can derive its entropy from the initial task distribution $p_0(\tau)$ and normalizing flows $\mathcal{G}=\{g_i\}_{i=1}^{M}$:
    \begin{equation}
        \begin{split}
            \mathbb{H}\Big[p_{\bm\phi_*}(\tau)\Big]
            =\mathbb{H}\Big[p_{0}(\tau)\Big]
            +\int p_{0}(\tau)\left[\sum_{i=1}^{M}\ln\left|\det\frac{\partial g_{i}}{\partial \tau^{i-1}}\right|\right]d\tau.
        \end{split}
    \end{equation}
\end{remark}
The above implies that the generated task distribution entropy is governed by the change of task identifiers in the probability measure of the task space.

\subsection{Strategies for Finding Equilibrium}
Given the previously formulated optimization objective, we propose to approach it with the help of estimated stochastic gradients.
As noticed, the involvement of adaptive expectation term $p_{\bm\phi}(\tau)$ requires extra considerations in optimization.

\textbf{Best Response Approximation.}
Given two players with completely distinguished purposes, the commonly used strategy to compute the equilibrium is the Best Response (BR), which means:
\begin{subequations}
    \begin{align}
        \bm\theta_{t+1}=\arg\min_{\bm\theta\in\bm\Theta}\mathcal{J}(\bm\theta,\bm\phi_t)\\
        \bm\phi_{t+1}=\arg\max_{\bm\phi\in\bm\Phi}\mathcal{J}(\bm\theta_{t+1},\bm\phi).
    \end{align}
\end{subequations}
 
For implementation convenience, we instead apply the gradient updates to the meta player and the distribution adversary, namely stochastic alternating gradient descent ascent (GDA).
The operations are entangled and result in the following iterative equations with the index $t$:
\begin{subequations}
    \begin{align}
        \bm\theta_{t+1}\leftarrow\bm\theta_{t}-\gamma_{1}
        \nabla_{\bm\theta}\mathcal{J}(\bm\theta_{t},\bm\phi_{t})
        \\
        \bm\phi_{t+1}\leftarrow\bm\phi_{t}+\gamma_{2}
        \nabla_{\bm\phi}\mathcal{J}(\bm\theta_{t+1},\bm\phi_{t}).
    \end{align}
    \label{stochastic_grad_opt}
\end{subequations}
This can be viewed as the gradient approximation for the BR strategy, which leads to at least a local Stackelberg equilibrium for the considered minimax problem \citep{jin2019minmax}.

\textbf{Stochastic Gradient Estimates \& Variance Reduction.}
Addressing the game-theoretic problem is non-trivial especially when it relates to distributions.
A commonly-used method is to perform the sample average approximation \textit{w.r.t.} Eq. (\ref{stochastic_grad_opt}).
It iteratively updates the parameters of the meta player and the distribution adversary to approximate the saddle point.

More specifically, we can have the Monte Carlo estimates of the stochastic gradients for the leader $\mathcal{P}_1$:
\begin{equation}
    \begin{split}
        \nabla_{\bm\theta}\mathcal{J}(\bm\theta,\bm\phi)
        =\int p_{\bm\phi}(\tau)
        \nabla_{\bm\theta}\mathcal{L}(\mathcal{D}_{\tau}^{Q},\mathcal{D}_{\tau}^{S};\bm\theta)d\tau
        \\
        \approx\frac{1}{K}\sum_{k=1}^{K}
        \nabla_{\bm\theta}\mathcal{L}(D_{\tau_k}^{Q},D_{\tau_k}^{S};\bm\theta).
    \end{split}
    \label{meta_player_obj}
\end{equation}
The form of stochastic gradients \textit{w.r.t.} the meta player parameter $\bm\theta$ is the meta-learning algorithm specific or model dependent.
We refer the reader to Algorithm \ref{armaml_pseudo}/\ref{arcnp_pseudo} as examples.

Now, we can derive the estimates with the help of REINFORCE algorithm \citep{williams1992simple} for the follower $\mathcal{P}_2$ and obtain the score function as:
\begin{equation}
    \begin{split}
        \nabla_{\bm\phi}\mathcal{J}(\bm\theta,\bm\phi)
        \approx\frac{1}{K}\sum_{k=1}^{K}
        \mathcal{L}(D_{\tau_k}^{Q},D_{\tau_k}^{S};\bm\theta)\nabla_{\bm\phi}\ln p_{\bm\phi}(\tau_k)\\
        +\frac{\lambda}{K}\sum_{k=1}^{K}\nabla_{\bm\phi}\ln p_{\bm\phi}(\tau_{k}^{-M}),
    \end{split}
    \label{meta_adversary_obj}
\end{equation}
where the particle $\tau_k\sim p_{\bm\phi}(\tau)$ denotes the task sampled from the generative task distribution, and $\tau_{k}^{-M}$ means the particle sampled from the initial task distribution to enable $\texttt{NN}_{\bm\phi}(\tau_{k}^{-M})=\tau_{k}$.

As validated in \citep{fu2006gradient}, the score estimator is an unbiased estimate of $\nabla_{\bm\phi}\mathcal{J}(\bm\theta,\bm\phi)$.
However, such a gradient estimator in Eq. (\ref{meta_adversary_obj}) mostly exhibits higher variances, which weakens the stability of training processes.
To reduce the variances, we utilize the commonly-used trick by including a constant baseline $\mathcal{V}=\mathbb{E}_{p_{\bm\phi}(\tau)}\left[\mathcal{L}(\mathcal{D}_{\tau}^{Q},\mathcal{D}_{\tau}^{S};\bm\theta)\right]\approx\frac{1}{K}\sum_{k=1}^K\mathcal{L}(D_{\tau_k}^{Q},D_{\tau_k}^{S};\bm\theta)$ for the score function, which results in:
\begin{equation}
    \begin{split}
        \nabla_{\bm\phi}\mathcal{J}(\bm\theta,\bm\phi):\approx
\frac{1}{K}\sum_{k=1}^{K}
        [\mathcal{L}(D_{\tau_k}^{Q},D_{\tau_k}^{S};\bm\theta)-\mathcal{V}]\nabla_{\bm\phi}\ln p_{\bm\phi}(\tau_k)\\
        +\frac{\lambda}{K}\sum_{k=1}^{K}\nabla_{\bm\phi}\ln p_{\bm\phi}(\tau_{k}^{-M}).
    \end{split}
    \label{meta_adversary_obj_1}
\end{equation}

Particularly, since the normalizing flow works as the distribution transformation in this work, please refer to Eq. (\ref{nf_logdeterm}) to obtain the derivative of the log-likelihood of the transformed task $\ln p_{\bm\phi}(\tau)$ \textit{w.r.t.} $\bm\phi$ inside Eq. (\ref{meta_adversary_obj_1}).
For easier analysis, we characterize the iteration sequence in optimization as
$\begin{bmatrix}
    \bm\theta_0\\
    \bm\phi_0
\end{bmatrix}
\mapsto
\cdots
\mapsto
\begin{bmatrix}
    \bm\theta_t\\
    \bm\phi_t
\end{bmatrix}
\mapsto
\begin{bmatrix}
    \bm\theta_{t+1}\\
    \bm\phi_{t+1}
\end{bmatrix}
\mapsto\cdots.
$

\begin{remark}[Solution as a Fixed Point]\label{exist_station_point}
The alternating GDA for solving Eq. (\ref{game_meta_obj}) results in the fixed point when $\begin{bmatrix}
    \bm\theta_{H+1}\\
    \bm\phi_{H+1}
\end{bmatrix}
=
\begin{bmatrix}
    \bm\theta_{H}\\
    \bm\phi_{H}
\end{bmatrix}$, or in other words $\begin{bmatrix}
    \bm\theta_{H}\\
    \bm\phi_{H}
\end{bmatrix}$ is stationary $\nabla\mathcal{J}(\bm\theta_H,\bm\phi_H)=0$.
\end{remark}

\begin{figure*}[h]
\begin{center}
\centerline{\includegraphics[width=0.95\textwidth]{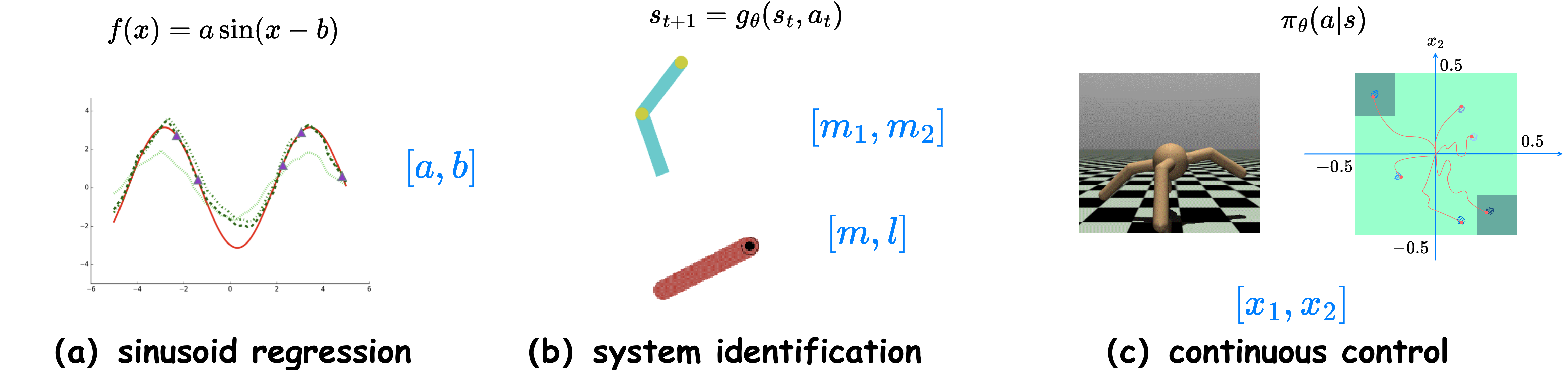}}
\caption{\textbf{Some Benchmarks in Evaluation.}
Blue-marked variables in the illustration denote task identifiers that guide the configuration of a specific task.
We place distributions over these task identifiers in generating diverse tasks for meta-learning.
}
\label{vis_benchmarks}
\end{center}
\end{figure*}

\subsection{Theoretical Analysis}

Built on the deduction of the local Stackelberg equilibrium's existence and the \textbf{Remark} \ref{exist_station_point}, we further perform analysis on the considered equilibrium 
$\begin{bmatrix}
    \bm\theta_{*}\\
    \bm\phi_{*}
\end{bmatrix}$, in terms of learning dynamics using the alternating GDA.
For notation simplicity, we denote the block terms inside the Hessian matrix $\mathbf{H}_{*}:=\nabla^2\mathcal{J}(\bm\theta_*,\bm\phi_*)$ around $[\bm\theta_*,\bm\phi_*]^T$ as $\begin{bmatrix}
\nabla^2_{\bm\theta\bm\theta}\mathcal{J} &\nabla^2_{\bm\theta\bm\phi}\mathcal{J} \\
\nabla^2_{\bm\phi\bm\theta}\mathcal{J} &\nabla^2_{\bm\phi\bm\phi}\mathcal{J} 
\end{bmatrix}\Big|_{[\bm\theta_*,\bm\phi_*]^T}
:=
\begin{bmatrix}
\mathbf{A} &\mathbf{B} \\
\mathbf{B}^T &\mathbf{C} 
\end{bmatrix}$.

\begin{theorem}[Convergence Guarantee]\label{converg_guarantee}
    Suppose that the \textbf{Assumption} \ref{assum_lipschitz} and the function condition of the (local) Stackelberg equilibrium $\Delta(\mathbf{A},\mathbf{B},\mathbf{C},\gamma_1,\gamma_2)<\frac{1}{2}$ are satisfied, where norms of the corresponding matrix are involved. 
    Then the following statements hold:
    \begin{enumerate}
        \item 
        The resulting iterated parameters $\{\cdots
        \mapsto
        [\bm\theta_t,\bm\phi_t]^T
        \mapsto
        [\bm\theta_{t+1},\\
        \bm\phi_{t+1}]^T
        \mapsto\cdots\}$ are Cauchy sequences;
        \item
        The optimization can guarantee at least the linear convergence to the local Stackelberg equilibrium with the rate $\sqrt{\Delta}$.
    \end{enumerate}
\end{theorem}

The \textbf{Theorem} \ref{converg_guarantee} clarifies learning rates $\gamma_1$ and $\gamma_2$'s influence on convergence and the required second-order derivative conditions of the resulting stationary point $[\bm\theta_*,\bm\phi_*]^T$.
And when the game arrives at convergence, the local Stackelberg equilibrium is the best response to these two players, which is at least a local min-max solution to Eq. (\ref{game_meta_obj}).

Next, we estimate the generalization bound of meta learners when confronting the generated task distribution shifts.
\begin{theorem}[Generalization Bound with the Distribution Adversary]\label{main_general_bound}
    Given the pretrained normalizing flows $\{g_i\}_{i=1}^M$, where $g_i$ is $(\ell_{a}, \ell_{b})$-bi-Lipschitz, and the pretrained meta learner $\bm\theta_*\in\bm\Theta$, we can derive the generalization bound with the initial task distribution $p$ uniform:
    \begin{equation}
        \begin{split}
            R_p^{\omega}(\bm\theta_*)
            \leq
            \hat{R}_p^{\omega}(\bm\theta_*)
            +\Upsilon(\mathcal{T})\left(\frac{\mathcal{C}\ln\frac{2Ke}{\mathcal{C}}+\ln\frac{4}{\delta}}{K}\right)^{\frac{3}{8}},
        \end{split}
    \end{equation}
where $\mathcal{C}=\text{Pdim}(\{\mathcal{L}(\cdot;\bm\theta):\bm\theta\in\bm\Theta\})$ denotes the pseudo-dimension in \citep{pollard1984convergence}, $ R_p^{\omega}(\bm\theta_*)$ and $\hat{R}_p^{\omega}(\bm\theta_*)$ are expected and empirical risks.
\end{theorem}
We refer the reader to Appendix \ref{append_sec:generalization} for formal \textbf{Theorem} \ref{main_general_bound} and proofs.
It reveals the connection between the bound and task complexity $\Upsilon(\mathcal{T})$, and more training tasks from initial distributions decrease the generalization error in adversarially distribution shifts.

\setlength{\tabcolsep}{3.0pt}
\begin{table*}[ht]
\caption{\textbf{Average mean square errors in 5-shot sinusoid regression/10-shot Acrobot system identification/10-shot Pendulum system identification with reported standard deviations (5 runs).} With $\alpha=0.5$, the best results are in pink (the lower, the better). U/N in benchmarks denote Uniform/Normal as the initial distribution type.}
\centering
\begin{adjustbox}{max width = 1.0\linewidth}
\begin{tabular}{|l|c|ccccc|ccccc|}
\toprule
\multirow{2}{*}{\textbf{Benchmark}} &\textbf{Meta-Test} &\multicolumn{5}{c|}{\texttt{\textbf{Average}}} & \multicolumn{5}{c|}{\texttt{\textbf{CVaR}}} \\
&\textbf{Distribution} &MAML &TR-MAML &DR-MAML &DRO-MAML &AR-MAML &MAML &TR-MAML &DR-MAML &DRO-MAML &AR-MAML \\
\bottomrule
\multirow{2}{*}{Sinusoid-\textbf{\texttt{U}}}
& {Initial}
&0.499\scriptsize{$\pm$0.01} 
&0.539\scriptsize{$\pm$0.01}
&0.479\scriptsize{$\pm$0.01}
&0.481\scriptsize{$\pm$0.01}
&\cellcolor{babypink}0.459\scriptsize{$\pm$0.01}
&0.858\scriptsize{$\pm$0.01} 
&0.868\scriptsize{$\pm$0.02}
&0.793\scriptsize{$\pm$0.02}
&0.816\scriptsize{$\pm$0.02}
&\cellcolor{babypink}0.782\scriptsize{$\pm$0.03}
\\
& {Adversarial}
&0.508\scriptsize{$\pm$0.01} 
&0.548\scriptsize{$\pm$0.01}
&0.499\scriptsize{$\pm$0.01}
&0.502\scriptsize{$\pm$0.02}
&\cellcolor{babypink}0.405\scriptsize{$\pm$0.01}
&0.883\scriptsize{$\pm$0.02} 
&0.879\scriptsize{$\pm$0.02}
&0.836\scriptsize{$\pm$0.01}
&0.826\scriptsize{$\pm$0.03}
&\cellcolor{babypink}0.671\scriptsize{$\pm$0.01}
\\
\cmidrule{2-12}
\multirow{2}{*}{Sinusoid-\textbf{\texttt{N}}}
& {Initial}
&0.578\scriptsize{$\pm$0.03} 
&0.628\scriptsize{$\pm$0.01}
&0.556\scriptsize{$\pm$0.01}
&0.562\scriptsize{$\pm$0.02}
&\cellcolor{babypink}0.554\scriptsize{$\pm$0.02}
&1.017\scriptsize{$\pm$0.05} 
&1.017\scriptsize{$\pm$0.02}
&\cellcolor{babypink}0.932\scriptsize{$\pm$0.02}
&0.983\scriptsize{$\pm$0.03}
&0.947\scriptsize{$\pm$0.03}
\\
& {Adversarial}
&0.496\scriptsize{$\pm$0.01} 
&0.511\scriptsize{$\pm$0.01}
&0.492\scriptsize{$\pm$0.02}
&0.493\scriptsize{$\pm$0.01}
&\cellcolor{babypink}0.404\scriptsize{$\pm$0.02}
&0.838\scriptsize{$\pm$0.03} 
&0.827\scriptsize{$\pm$0.02}
&0.807\scriptsize{$\pm$0.03}
&0.835\scriptsize{$\pm$0.01}
&\cellcolor{babypink}0.672\scriptsize{$\pm$0.03}
\\
\midrule
\multirow{2}{*}{Acrobot-\textbf{\texttt{U}}}
& {Initial}
&0.244\scriptsize{$\pm$0.01}
&0.233\scriptsize{$\pm$0.00}
&0.222\scriptsize{$\pm$0.00}
&0.237\scriptsize{$\pm$0.00}
&\cellcolor{babypink}0.219\scriptsize{$\pm$0.01}
&0.336\scriptsize{$\pm$0.01}
&0.320\scriptsize{$\pm$0.00}
&0.303\scriptsize{$\pm$0.00}
&0.322\scriptsize{$\pm$0.01}
&\cellcolor{babypink}0.298\scriptsize{$\pm$0.00}
\\
& {Adversarial}
&0.243\scriptsize{$\pm$0.00} 
&0.238\scriptsize{$\pm$0.01}
&0.235\scriptsize{$\pm$0.01}
&0.244\scriptsize{$\pm$0.00}
&\cellcolor{babypink}0.230\scriptsize{$\pm$0.00}
&0.341\scriptsize{$\pm$0.01} 
&0.320\scriptsize{$\pm$0.01}
&0.325\scriptsize{$\pm$0.01}
&0.333\scriptsize{$\pm$0.01}
&\cellcolor{babypink}0.306\scriptsize{$\pm$0.01}
\\
\cmidrule{2-12}
\multirow{2}{*}{Acrobot-\textbf{\texttt{N}}}
& {Initial}
&0.231\scriptsize{$\pm$0.00} 
&0.225\scriptsize{$\pm$0.00}
&0.227\scriptsize{$\pm$0.00}
&0.222\scriptsize{$\pm$0.00}
&\cellcolor{babypink}0.215\scriptsize{$\pm$0.00}
&0.321\scriptsize{$\pm$0.01} 
&0.311\scriptsize{$\pm$0.00}
&0.316\scriptsize{$\pm$0.01}
&0.309\scriptsize{$\pm$0.01}
&\cellcolor{babypink}0.301\scriptsize{$\pm$0.01}
\\
& {Adversarial}
&0.246\scriptsize{$\pm$0.00} 
&0.237\scriptsize{$\pm$0.00}
&0.241\scriptsize{$\pm$0.00}
&0.242\scriptsize{$\pm$0.00}
&\cellcolor{babypink}0.229\scriptsize{$\pm$0.00}
&0.338\scriptsize{$\pm$0.00} 
&0.327\scriptsize{$\pm$0.01}
&0.327\scriptsize{$\pm$0.00}
&0.332\scriptsize{$\pm$0.01}
&\cellcolor{babypink}0.314\scriptsize{$\pm$0.01}
\\
\midrule
\multirow{2}{*}{Pendulum-\textbf{\texttt{U}}}
& {Initial}
&0.648\scriptsize{$\pm$0.02}
&0.694\scriptsize{$\pm$0.01}
&0.634\scriptsize{$\pm$0.01}
&0.630\scriptsize{$\pm$0.02}
&\cellcolor{babypink}0.627\scriptsize{$\pm$0.01}
&0.799\scriptsize{$\pm$0.03}
&0.780\scriptsize{$\pm$0.02}
&0.744\scriptsize{$\pm$0.01}
&0.751\scriptsize{$\pm$0.03}
&\cellcolor{babypink}0.733\scriptsize{$\pm$0.02}
\\
& {Adversarial}
&0.672\scriptsize{$\pm$0.01} 
&0.724\scriptsize{$\pm$0.01}
&0.669\scriptsize{$\pm$0.01}
&0.674\scriptsize{$\pm$0.00}
&\cellcolor{babypink}0.660\scriptsize{$\pm$0.01}
&0.845\scriptsize{$\pm$0.02} 
&0.854\scriptsize{$\pm$0.02}
&0.808\scriptsize{$\pm$0.02}
&0.826\scriptsize{$\pm$0.01}
&\cellcolor{babypink}0.778\scriptsize{0.01}\\
\cmidrule{2-12}
\multirow{2}{*}{Pendulum-\textbf{\texttt{N}}}
& {Initial}
&0.596\scriptsize{$\pm$0.00} 
&0.637\scriptsize{$\pm$0.01}
&\cellcolor{babypink}0.574\scriptsize{$\pm$0.01}
&0.582\scriptsize{$\pm$0.00}
&0.586\scriptsize{$\pm$0.01}
&0.715\scriptsize{$\pm$0.01} 
&0.720\scriptsize{$\pm$0.01}
&\cellcolor{babypink}0.685\scriptsize{$\pm$0.01}
&0.695\scriptsize{$\pm$0.01}
&0.694\scriptsize{$\pm$0.01}
\\
& {Adversarial}
&0.664\scriptsize{$\pm$0.02} 
&0.702\scriptsize{$\pm$0.01}
&0.660\scriptsize{$\pm$0.02}
&0.677\scriptsize{$\pm$0.02}
&\cellcolor{babypink}0.635\scriptsize{$\pm$0.01}
&0.861\scriptsize{$\pm$0.03} 
&0.837\scriptsize{$\pm$0.02}
&0.817\scriptsize{$\pm$0.03}
&0.860\scriptsize{$\pm$0.04}
&\cellcolor{babypink}0.777\scriptsize{$\pm$0.03}
\\
\bottomrule
\end{tabular}
\end{adjustbox}
\label{test_regression}
\end{table*}

\section{Experiments}\label{exp_sec}

Previous sections recast the adversarially task robust meta-learning to a Stackelberg game, specify the equilibrium, and analyze theoretical properties in distribution generation.
This section focuses on the evaluation, and baselines constructed from typical risk minimization principles are reported in \textbf{Table} \ref{meta_learning_principles}.
These include vanilla MAML \citep{finn2017model}, DRO-MAML \citep{sagawa2019distributionally}, TR-MAML \citep{collins2020task}, DR-MAML \citep{wang2023sim}, and AR-MAML (ours).

Technically, we mainly answer the following \textbf{Research Questions} (\textbf{RQ}s):
\begin{enumerate}
    \item\textit{Does adversarial training help improve few-shot adaptation robustness in case of task distribution shifts?}

    \item\textit{How does the type of the initial task distribution influence the performance of resulting solutions?}

    \item\textit{Can generative modeling the task distribution discover meaningful task structures and afford interpretability?}
\end{enumerate}

\textit{Implementation \& Examination Setup.}
As our approach is agnostic to meta-learning methods, we mainly employ AR-MAML as the implementation of this work.
Concerning the meta testing distribution, tasks are from the initial task distribution and the adversarial task distribution, respectively.
The latter corresponds to the generated task distribution under shift constraints after convergence.

\textit{Evaluation Metrics.} Here, we use both the average risk and conditional value at risk ($\text{CVaR}_{\alpha}$) in evaluation metrics, where $\text{CVaR}_{\alpha}$ can be viewed as the worst group performance in \citep{sagawa2019distributionally}.

\subsection{Benchmarks}
We consider the few-shot synthetic regression, system identification, and meta reinforcement learning to test fast adaptation robustness with typical baselines.
Notably, the task is specified by the generated task identifiers as shown in Figure \ref{vis_benchmarks}.

\textit{Synthetic Regression.}
The same as that in \citep{finn2017model}, we conduct experiments in sinusoid functions.
The goal is to uncover the function $f(x)=a\sin(x-b)$ with $K$-shot randomly sampled function points.
And the task identifiers are the amplitude $a$ and phase $b$.

\textit{System Identification.}
Here, we take the Acrobot System \citep{sutton1998introduction} and the Pendulum System \citep{lee2020context} to perform system identification.
In the Acrobot System, we generate different dynamical systems as tasks by varying masses of two pendulums.
And the task identifiers are the pendulum mass parameters $m_1$ and $m_2$.
In the Pendulum System, the system dynamics are distinguished by varying the mass and the length of the pendulum.
And the task identifiers are the mass parameter $m$ and the length parameter $l$.
For both benchmarks, we collect the dataset of state transitions with a complete random policy to interact with sampled environments.
The goal is to predict state transitions conditioned on randomly sampled context transitions from an unknown dynamical system.

\textit{Meta Reinforcement Learning.}
We evaluate the role of task distributions in meta-learning continuous control.
In detail, the Point Robot in \citep{finn2017model} and the Ant-Pos Robot in Mujoco \citep{todorov2012mujoco} are included as navigation environments.
We respectively vary goal/position locations as task identifiers within a designed range to generate diverse tasks.
The goal is to seek a policy that guides the robot to the target location with a few episodes derived from an environment.

We refer the reader to Appendix \ref{append:exp setup} for set-ups, hyper-parameter configurations and additional experimental results.

\subsection{Empirical Result Analysis}
Here, we report the experimental results, perform analysis and answer the raised \textbf{RQ}s (1)/(2).

\begin{figure}[h]
\begin{center}
\centerline{\includegraphics[width=0.47\textwidth]{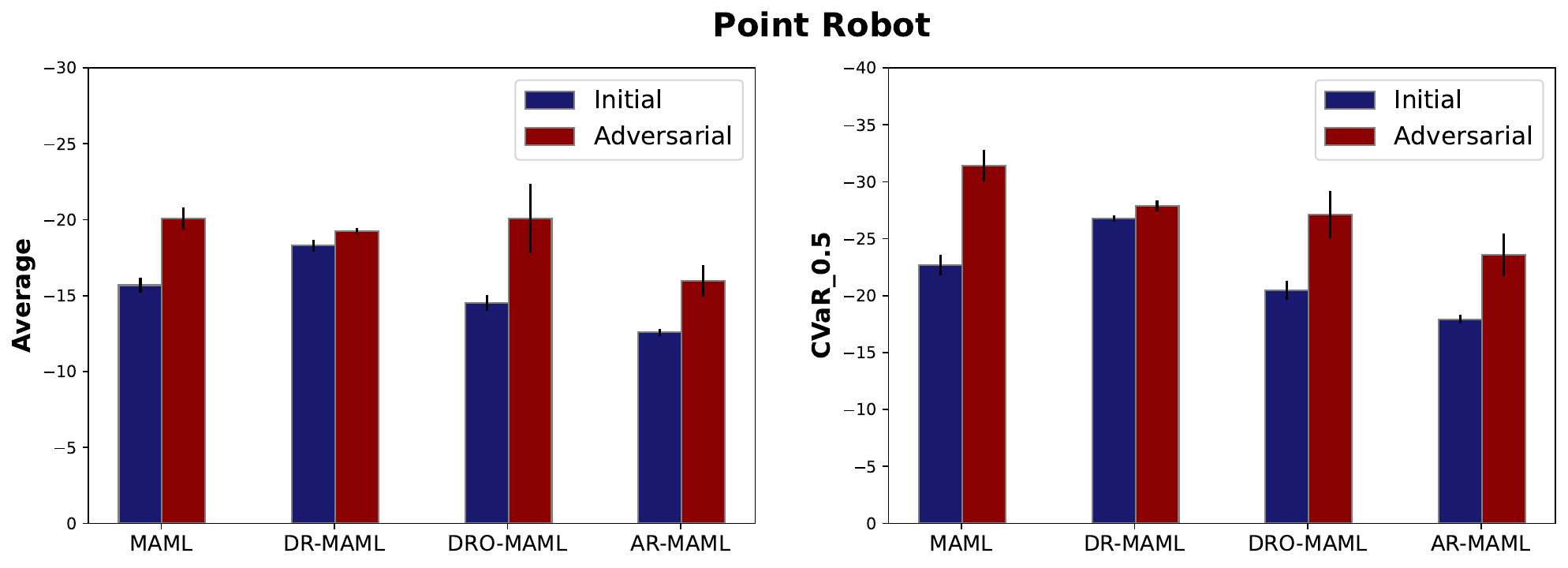}}
\caption{\textbf{Meta Testing Returns in Point Robot Navigation Tasks (4 runs).} 
The charts report average and $\text{CVaR}_{\alpha}$ returns with $\alpha=0.5$ in initial and adversarial distributions, with standard error bars indicated by black vertical lines. The higher, the better.
}
\label{point_robot}
\end{center}
\vspace{-15pt}
\end{figure}

\begin{figure}[h]
\begin{center}
\centerline{\includegraphics[width=0.47\textwidth]{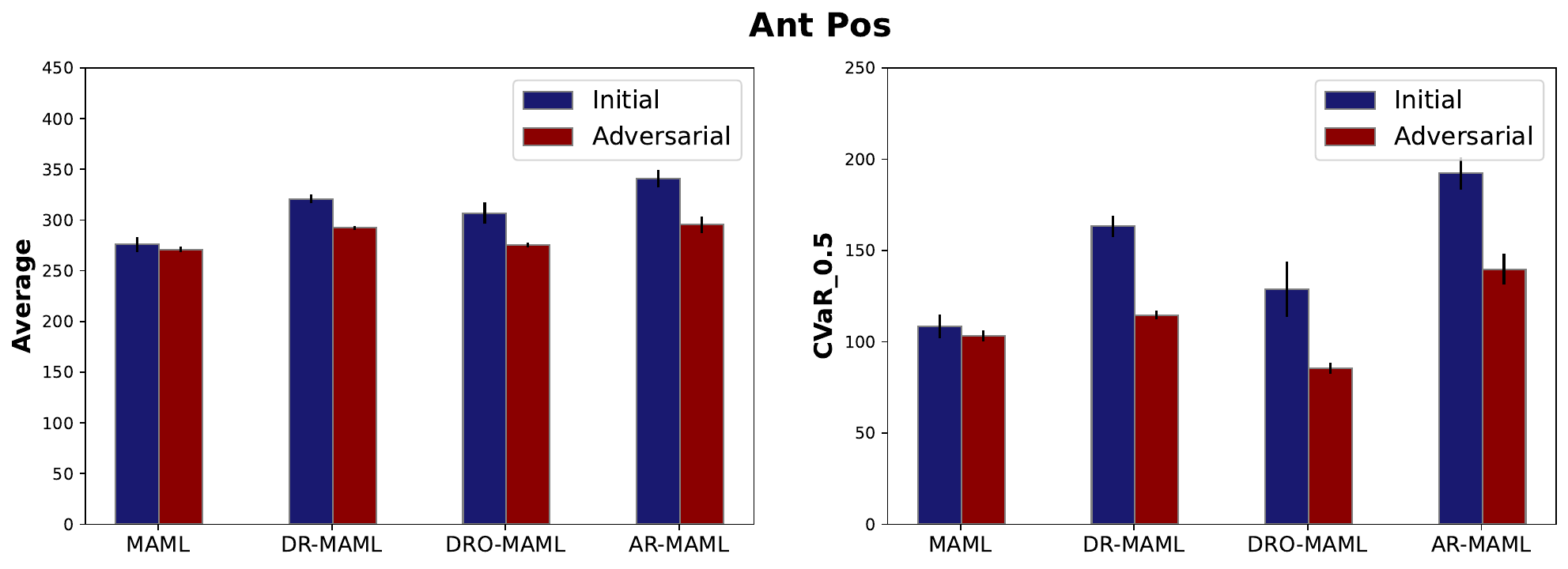}}
\caption{\textbf{Meta Testing Returns in Ant Pos Tasks (4 runs).}
The charts report average and $\text{CVaR}_{\alpha}$ returns with $\alpha=0.5$ in initial and adversarial distributions, with standard error bars indicated by black vertical lines. 
The higher, the better.
}
\label{ant_pos}
\end{center}
\vspace{-15pt}
\end{figure}

\textit{Overall Performance}:
Table \ref{test_regression} shows that
AR-MAML mostly outperforms others in the adversarial distribution, seldom sacrificing performance in the initial distribution.
Similar to observations in \citep{wang2023sim}, task distributionally robust optimization methods, like DR-MAML and DRO-MAML, not only retain robustness advantage on shifted distribution but also sometimes boost average performance on the initial distribution.
Cases with two types of initial task distributions (Uniform/Normal) come to similar conclusions on average and $\text{CVaR}_{\alpha}$ performance.
Figures \ref{point_robot}/\ref{ant_pos} show the meta reinforcement learning results for Point Robot and Ant Pos navigation tasks.
AR-MAML exhibits similar superiority on both continuous control benchmarks compared to baselines. 

\begin{figure}[h]
\begin{center}
\centerline{\includegraphics[width=0.46\textwidth]{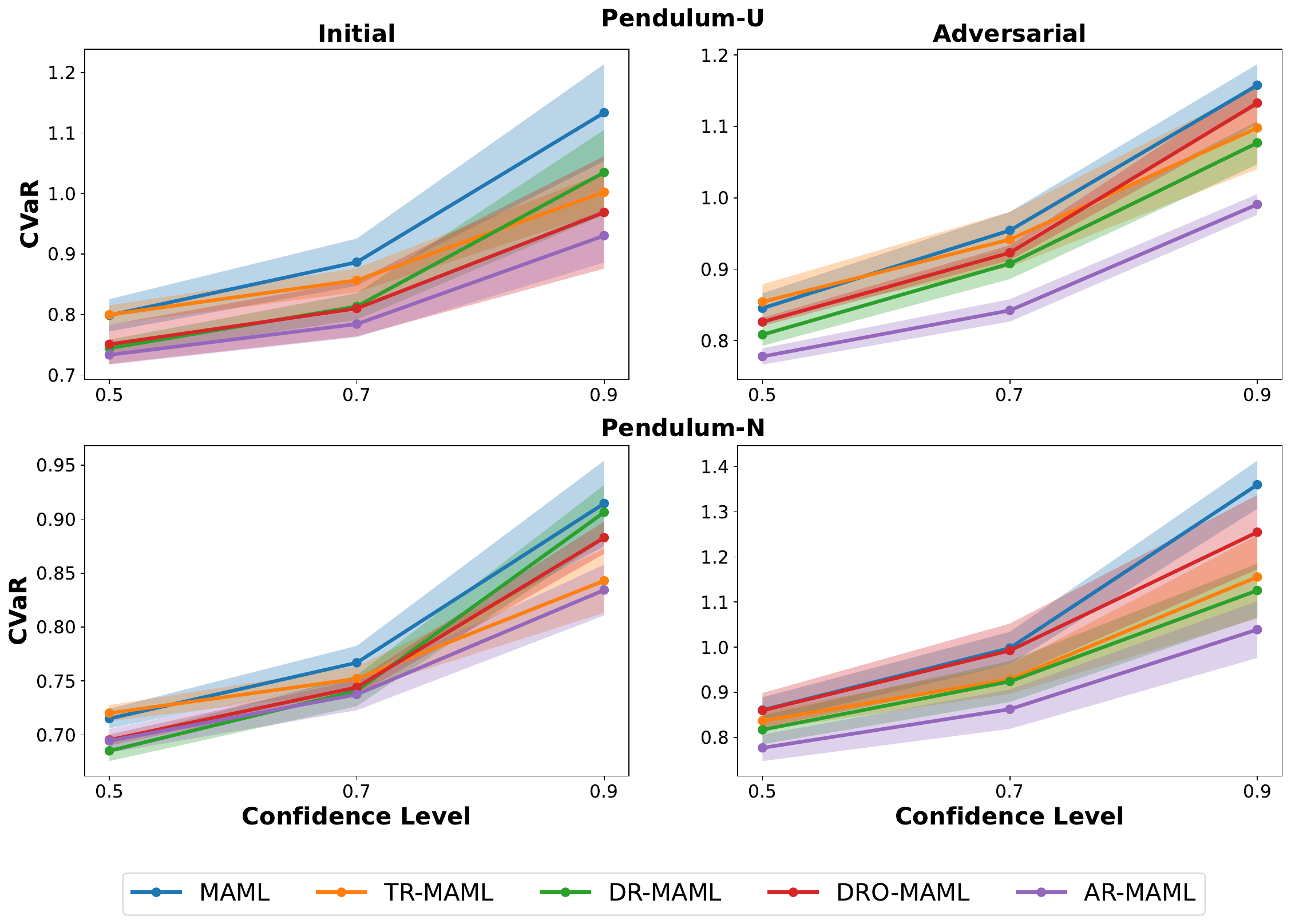}}
\caption{\textbf{$\text{CVaR}_{\alpha}$ MSEs with Various Confidence Level $\alpha$.}
Pendulum-U/N denotes Uniform/Normal as the initial distribution type.
The plots report meta testing $\text{CVaR}_{\alpha}$ MSEs in initial and adversarial distributions with standard error in shadow regions.
}
\label{pendulum_cvar}
\end{center}
\vspace{-15pt}
\end{figure}

\begin{figure}[h]
\begin{center}
\centerline{\includegraphics[width=0.47\textwidth]{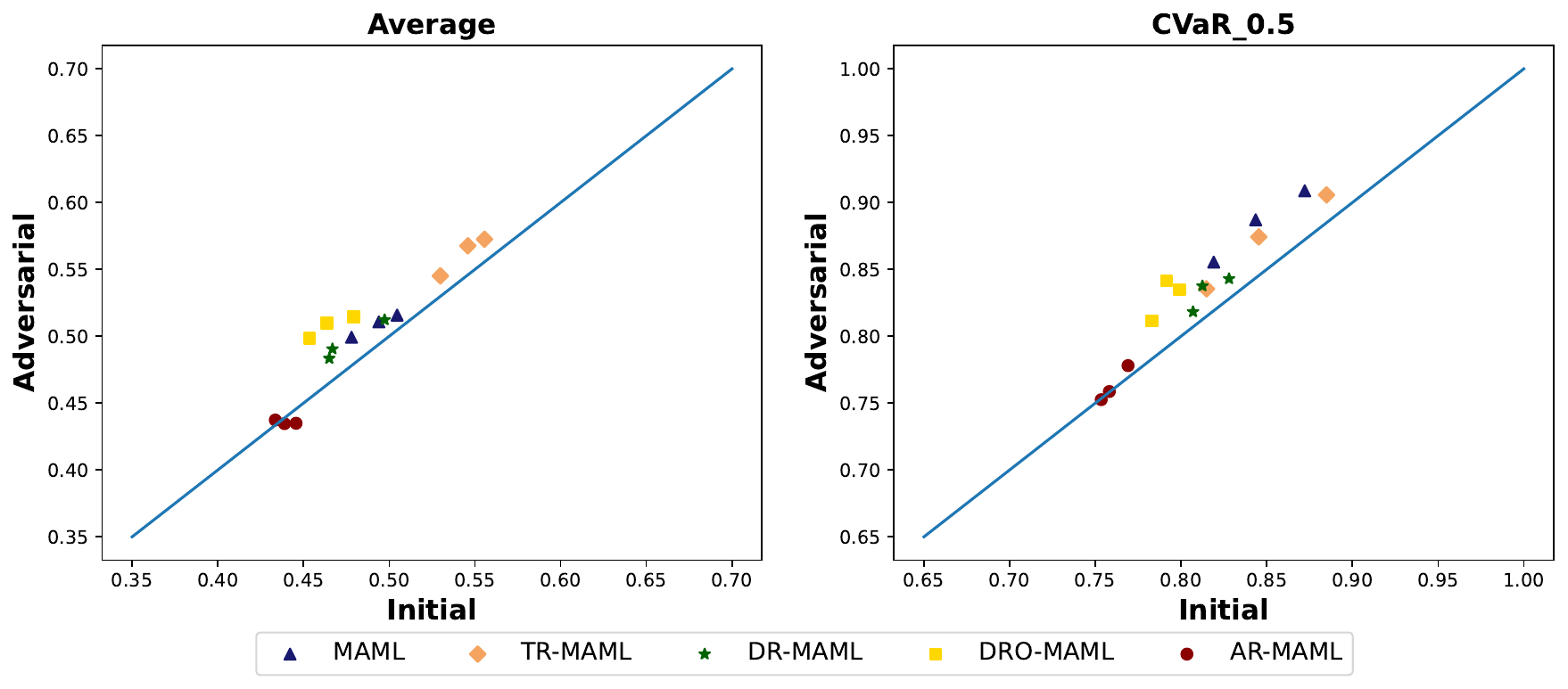}}
\caption{\textbf{Meta testing MSEs on the initial distribution (x-axis) and on the adversarial distribution (y-axis).}
The $y=x$ line serves as a baseline for comparison. Models above this line show increased losses when faced with distribution shifts, indicating a decline in performance compared to the standard test set.
}
\label{Sinusoid_scatter}
\end{center}
\vspace{-15pt}
\end{figure}

\textit{Multiple Tail Risk Robustness}:
Note that CVaR metrics imply the model's robustness under the subpopulation shift.
Figure \ref{pendulum_cvar} reports $\text{CVaR}_{\alpha}$ values with various confidence values on pendulum system identification.
The AR-MAML's merits in handling the proportional worst cases are consistent across diverse levels.
We also illustrate and include these statistics on other benchmarks in Appendix \ref{append: additional experimental results}.
Moreover, as suggested in \citep{taori2020measuring}, a robust learner seldom encounters a performance gap between a standard (initial) test set and a test set with a distribution shift (adversarial).
Figure \ref{Sinusoid_scatter} validates the meta-learners' robustness on sinusoid regression, where AR-MAML's results are more proximal to the $y=x$ line than other baselines.

\begin{figure}[h]
\begin{center}
\centerline{\includegraphics[width=0.47\textwidth]{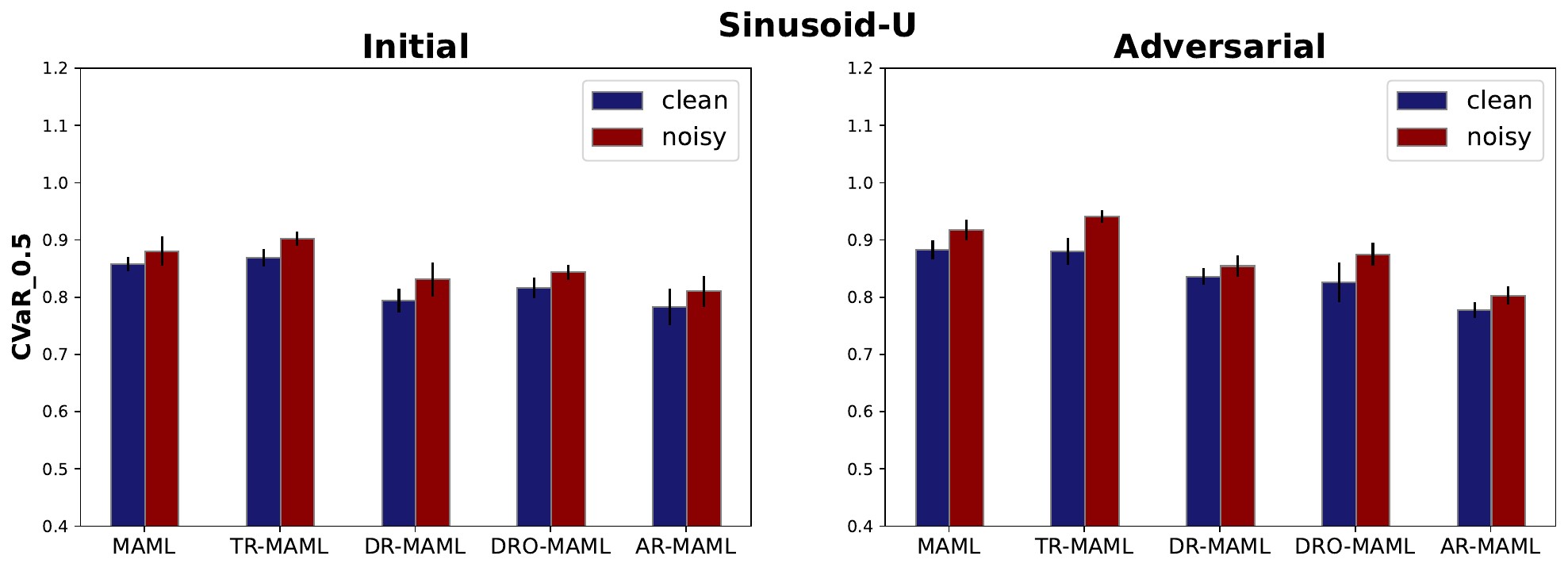}}
\caption{\textbf{Meta Testing Performance in Clean and Noisy Tasks.}
The noisy tasks are constructed by adding noise on the outputs of the support dataset.
Reported are testing $\text{CVaR}_{\alpha}$ MSEs with $\alpha=0.5$, where black vertical lines indicate standard error bars.
}
\label{sinusoid_noise_cvar}
\end{center}
\vspace{-15pt}
\end{figure}

\textit{Random Perturbation Robustness}:
We also test meta-learners' robustness to random noise from the support dataset.
To do so, we take sinusoid regression and inject random noise into the support set, i.e., the noise is drawn from a Gaussian distribution $\mathcal{N}(0,0.1^2)$ and added to the output $y$.
Figure \ref{sinusoid_noise_cvar} illustrates that AR-MAML's performance degradation is somewhat less than others on the adversarial distribution.
The noise exhibits similar effects on AR-MAML and DR-MAML on the initial distribution, harming performance severely.
AR-MAML and DR-MAML still exhibit lower MSEs than other baselines for all cases.
This indicates the adversarial training mechanism can also bring more robustness to challenging test scenarios with random noise.

\subsection{Task Structure Analysis}
In response to \textbf{RQ} (3), we turn to the analysis of the learned distribution adversary.
As a result, we visualize the adversarial task probability density.

\begin{figure}[h]
\begin{center}
\centerline{\includegraphics[width=0.5\textwidth]{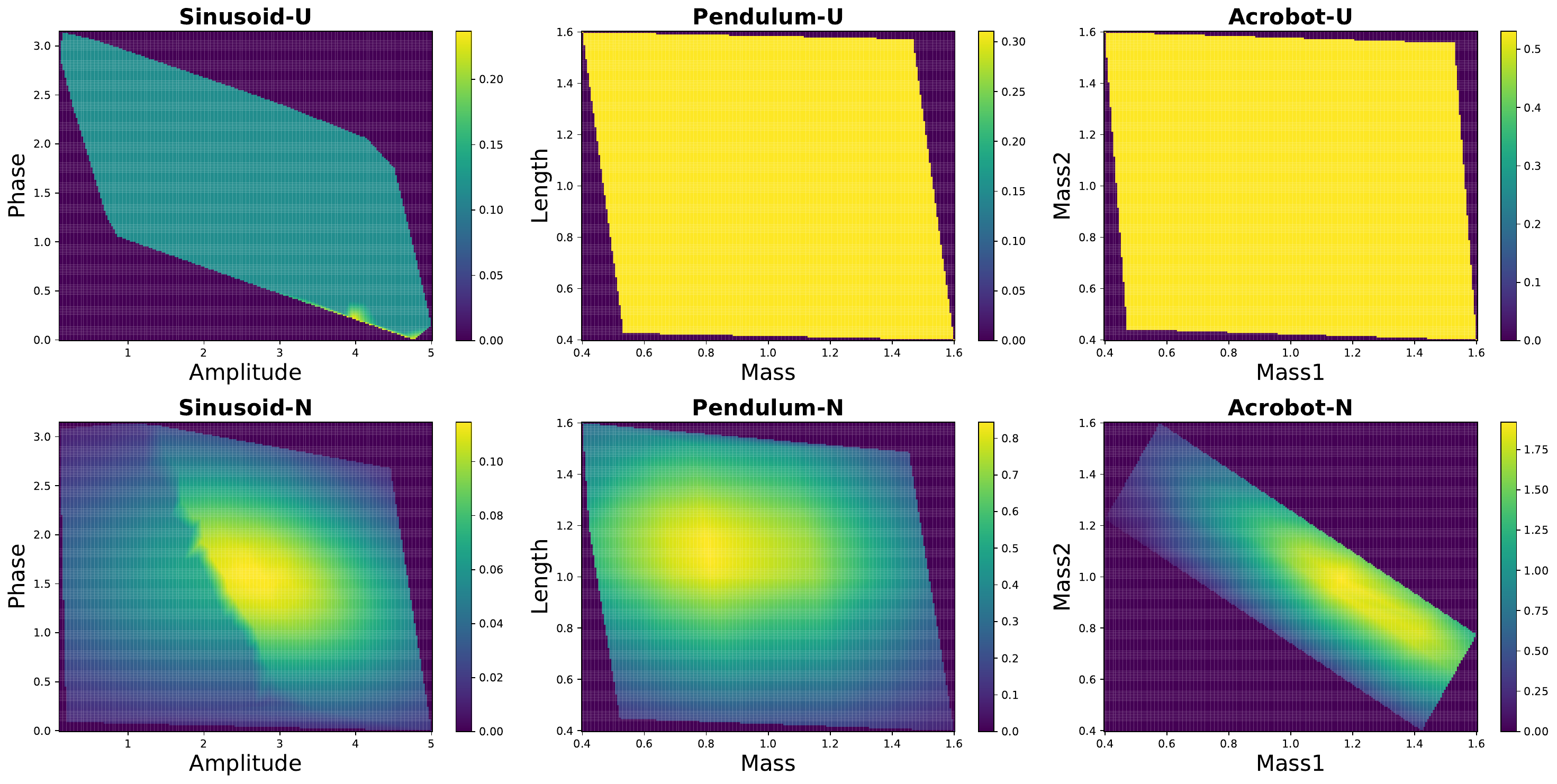}}
\caption{\textbf{Adversarial Task Probability Distribution.}
The plots show the adversarial distributions resulting from two different initial distributions: uniform (top row) and normal (bottom row).
}
\label{Task_structure}
\end{center}
\vspace{-15pt}
\end{figure}

\textit{Explicit Task Distribution}:
As displayed in Figure \ref{Task_structure}, our approach enables the discovery of explicit task structures regarding problem-solving.
The general learned patterns seem to be regardless of the initial task distributions.
In sinusoid regression, more probability mass is allocated in the region with $[3.0,5.0]\times[0.0,1.0]$, which reveals more difficulties in adaptation with larger amplitude descriptors.
For the Pendulum, the distribution adversary assigns less probability mass to two corner regions, implying that the combination of higher masses and longer pendulums or lower masses and shorter pendulums is easier to predict.
Similar phenomena are observed in mass combinations of Acrobat systems.
Consistently, the existence of constraint decreases all task distribution entropies to a certain level, which we report in Appendix \ref{append:exp setup}.
Though such a decrease brings more concentration on some task subsets, AR-MAML still probably fails to cover other challenging combinations in mode collapse.

\textit{Initial Task Distributions' Influence on Structures}:
Comparing the top and the bottom of Figure \ref{Task_structure}, we notice that the uniform and the normal initial distribution results in similar patterns after normalizing flows' transformations on separate benchmarks.
The normal initial distribution can be transformed into smooth ones and captures high-density regions around centroids.

\subsection{Other Investigations}
Here, we conduct additional investigations through the following perspectives.

\begin{figure}[h]
\begin{center}
\centerline{\includegraphics[width=0.47\textwidth]{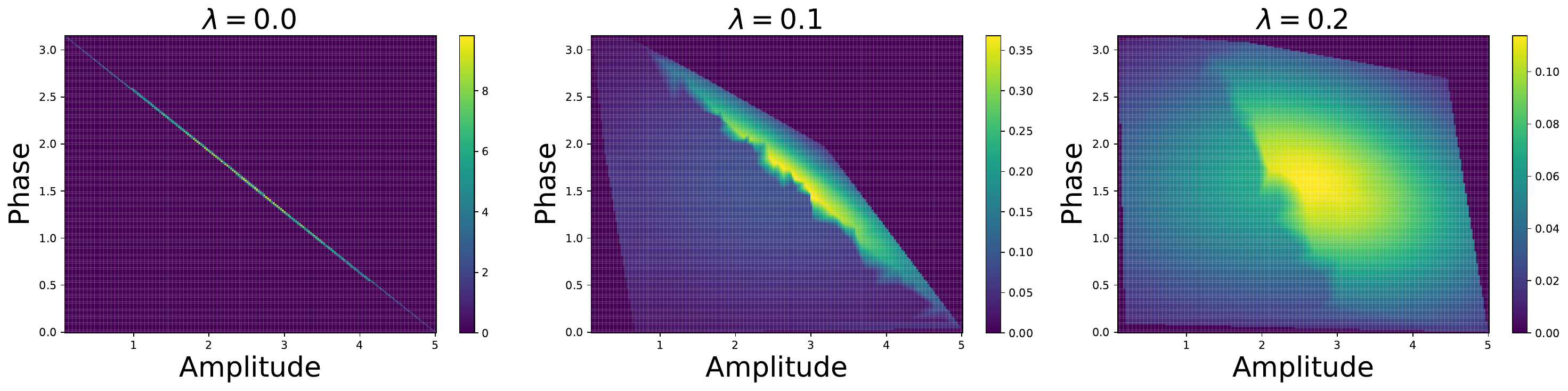}}
\caption{\textbf{Adversarial Task Probability Distribution on Sinusoid Regression with Various Lagrange Multipliers $\lambda$.}
}
\label{Lagrange_sinusoid_task_structure}
\end{center}
\vspace{-15pt}
\end{figure}

\begin{figure}[h]
\begin{center}
\centerline{\includegraphics[width=0.47\textwidth]{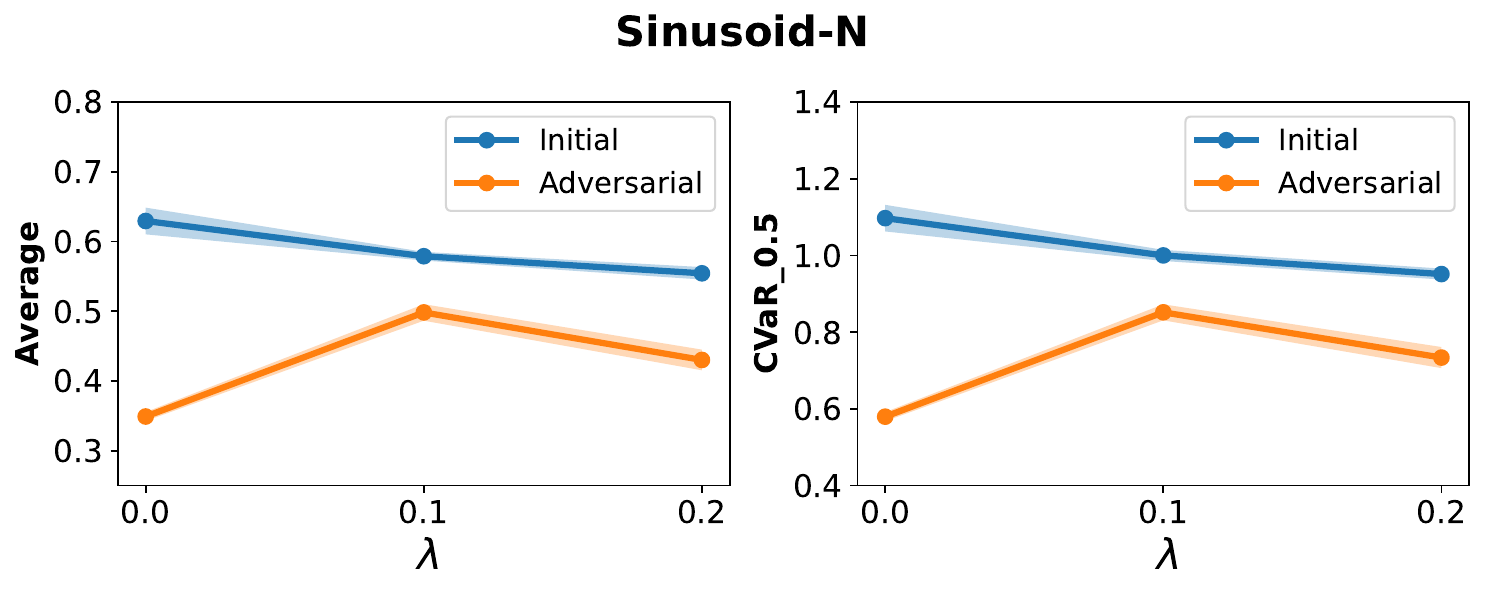}}
\caption{\textbf{Meta testing MSEs with various lagrange multiplier $\lambda$.}
Reported are testing average and $\text{CVaR}_{\alpha}$ MSEs with $\alpha=0.5$ with standard error in shadow regions.}
\label{Lagrange_Sinusoid-N}
\end{center}
\vspace{-15pt}
\end{figure}

\textit{Impacts of Shift Distribution Constraints:}
Our studied framework allows the task distribution to shift at a certain level.
In Eq. (\ref{main_obj}), larger $\lambda$ values tend to cause the generated distribution to collapse into the initial distribution.
Consequently, we empirically test the naive and severe adversarial training, e.g., setting $\lambda=\{0.0,0.1,0.2\}$ on sinusoid regression.
As displayed in Figure \ref{Lagrange_sinusoid_task_structure}, the generated distribution with $\lambda=0.0$ suffers from severe mode collapse, merely covering diagonal regions in the task space.
Such a curse is alleviated with increasing $\lambda$ values.
In Figure \ref{Lagrange_Sinusoid-N}, the meta learner, after heavy distribution shifts, catastrophically fails to generalize well in the initial distribution, illustrating higher adaptation risks in $\lambda=0.0$.

\textit{Compatibility with Other Meta-learning Methods:}
Besides the AR-MAML, we also check the effect of adversarially task robust training with other meta-learning methods.
Here, AR-CNP in Example \ref{example_arcnp} is employed in the evaluation.
Take the sinusoid regression as an example. 
Table \ref{CNP_test_results} observes comparable performance between AR-CNP and DR-CNP on the initial task distribution, while results on the adversarial task distribution uncover a significant advantage over others, particularly on robustness metrics, namely $\text{CVaR}_{\alpha}$ values.

\setlength{\tabcolsep}{6.0pt}
\begin{table}[ht]
\caption{\textbf{Meta testing MSEs in 5-shot sinusoid regression.} With $\alpha=0.5$, the best results are in pink (the lower, the better).}
\centering
\begin{adjustbox}{max width = 1.0\linewidth}
\begin{tabular}{l|cc|cc}
\toprule
& \multicolumn{2}{c|}{\texttt{\textbf{Average}}} & \multicolumn{2}{c}{\texttt{\textbf{CVaR}}} \\
Method & Initial & Adversarial & Initial & Adversarial \\
\bottomrule
CNP & 0.023\scriptsize{$\pm$0.001} 
&0.026\scriptsize{$\pm$0.004}
&0.041\scriptsize{$\pm$0.003}
&0.045\scriptsize{$\pm$0.007}
\\
TR-CNP & 0.048\scriptsize{$\pm$0.002} 
&0.050\scriptsize{$\pm$0.002}
&0.076\scriptsize{$\pm$0.004}
&0.079\scriptsize{$\pm$0.004}
\\
DR-CNP &0.021\scriptsize{$\pm$0.001} 
&0.023\scriptsize{$\pm$0.002}
&0.034\scriptsize{$\pm$0.003}
&0.037\scriptsize{$\pm$0.003}
\\
DRO-CNP & 0.023\scriptsize{$\pm$0.001} 
&0.025\scriptsize{$\pm$0.002}
&0.039\scriptsize{$\pm$0.003}
&0.041\scriptsize{$\pm$0.004}
\\
\bottomrule
AR-CNP(Ours) &\cellcolor{babypink} 0.019\scriptsize{$\pm$0.001}
&\cellcolor{babypink} 0.018\scriptsize{$\pm$0.002}
&\cellcolor{babypink} 0.033\scriptsize{$\pm$0.001}
&\cellcolor{babypink} 0.029\scriptsize{$\pm$0.003}
\\

\bottomrule
\end{tabular}
\end{adjustbox}
\label{CNP_test_results}
\end{table}

\section{Conclusions}\label{conclusion_sec}

\textbf{Discussions \& Society Impacts.} 
This work develops a game-theoretical approach for generating explicit task distributions in an adversarial way and contributes to theoretical understandings.
In extensive scenarios, our approach improves adaptation robustness in constrained distribution shifts and enables the discovery of interpretable task structures in optimization.

\noindent
\textbf{Limitations \& Future Work.}
The task distribution in this work relies on the task identifier, which can be inaccessible in some cases, e.g., few-shot classification.
Also, the adopted strategy to derive the game solution is approximate, leading to suboptimality in optimization. 
Hence, future efforts can be made to overcome these limitations and facilitate robust adaptation in applications. 

\begin{acks}
This work is funded by National Natural Science Foundation of China (NSFC) with the Number \# 62306326 and \# 62495091.
And we thank Dong Liang, Yuhang Jiang, Chen Chen, Daming Shi, other anonymous reviewers, and KDD2025 Area Chairs Prof. Yan Liu and Prof. Auroop R Ganguly for suggestions and helpful discussions.
\end{acks}

\bibliographystyle{ACM-Reference-Format}
\bibliography{kdd_2024}

\title{Supplementary Materials}

\newpage
\appendix
\onecolumn
\tableofcontents

\newpage
\section{Quick Guideline for This Work}

Here, we include some guidelines for this work.
Our focus is to explicitly generate the task distribution with adversarial training.
The use of normalizing flows enables task structure discovery under the risk minimization principle.
The theoretical understanding is from the Stackelberg game, together with some analyses.
Allowing the task distribution to shift at an acceptable level is a promising topic in meta-learning and can help improve robustness in the presence of worse cases.
The following further complements these points.

\subsection{Pseudo Code of the AR-MAML \& AR-CNP}
This section lists the pseudo-code for implementing AR-MAML and AR-CNPs in practice.
The difference between AR-MAML and AR-CNP lies in the inner loop, as CNP does not require additional gradient updates in evaluating the meta-learner.

\begin{algorithm}
\SetAlgoLined
\SetKwInOut{Input}{Input}
\SetKwInOut{Output}{Output}
\Input{Initial task distribution $p_{0}(\tau)$;
 Task batch size $K$; Update frequency $u$;
 Learning rates of players: $\{\gamma_{1,1},\gamma_{1,2},\gamma_2\}$.}
\Output{Meta-trained model parameter $\bm\theta$.}

Randomly initialize the model parameter $\bm\theta$;

Set the initial iteration number $t\leftarrow0$;

\While{not converged}{
Sample a batch of tasks $\{\tau_i\}_{i=1}^{K}\sim p_{\bm\phi}(\tau)$;

\tcp{\textcolor{blue}{\texttt{\textbf{Leader Inner Gradient Descent}}}}
\For{$i=1$ {\bfseries to} $K$}{
Compute the task-specific gradient: $\nabla_{\bm\theta}\mathcal{L}(\mathcal{D}_{\tau_i}^{S};\bm\theta)$;

Perform gradient updates as fast adaptation:

$\bm\theta_{i}\leftarrow\bm\theta-\gamma_{1,1}\nabla_{\bm\theta}\mathcal{L}(\mathcal{D}_{\tau_i}^{S};\bm\theta)$;
}

Update the iteration number: $t\leftarrow t+1$;

\tcp{\textcolor{blue}{\texttt{\textbf{Leader Outer Gradient Descent}}}}
Perform meta initialization updates:

\quad\quad$\bm\theta\leftarrow\bm\theta-\frac{\gamma_{1,2}}{K}\sum_{i=1}^{K}\nabla_{\bm\theta}\mathcal{L}(\mathcal{D}_{\tau_i}^{Q};\bm\theta_i)$;

\If{t\%u=0}{

\tcp{\textcolor{blue}{\texttt{\textbf{Follower Gradient Ascent}}}}

Compute the baseline for the follower:

$\mathcal{V}\approx\frac{1}{K}\sum_{k=1}^K\mathcal{L}(D_{\tau_k}^{Q};\bm\theta)$;

Perform task distribution $p_{\bm\phi}(\tau)$ updates:

$\bm\phi\leftarrow\bm\phi+\frac{\gamma_{2}}{K}\sum_{k=1}^{K}
        [\mathcal{L}(D_{\tau_k}^{Q};\bm\theta)-\mathcal{V}]\nabla_{\bm\phi}\ln p_{\bm\phi}(\tau_k)
        +\frac{\lambda\gamma_{2}}{K}\sum_{k=1}^{K}\nabla_{\bm\phi}\ln p_{\bm\phi}(\tau_{k}^{-M}).$

}}
\caption{Adversarially Task Robust MAML}
\label{armaml_pseudo}
\end{algorithm}

\begin{algorithm}[H]
\SetAlgoLined
\SetKwInOut{Input}{Input}
\SetKwInOut{Output}{Output}
\Input{Initial task distribution $p_{0}(\tau)$;
 Task batch size $K$; Update frequency $u$;
 Learning rates of players: $\{\gamma_{1},\gamma_2\}$.}
\Output{Meta-trained model parameter $\bm\theta=[\bm\theta_1,\bm\theta_2]$.}

Randomly initialize the model parameter $\bm\theta$;

Set the initial iteration number $t\leftarrow0$;

\While{not converged}{
Sample a batch of tasks $\{\tau_i\}_{i=1}^{K}\sim p_{\bm\phi}(\tau)$;

\tcp{\textcolor{blue}{\texttt{\textbf{Leader Gradient Descent}}}}
Compute the task-specific gradient for the leader: $\nabla_{\bm\theta}\mathcal{L}(\mathcal{D}_{\tau_i}^{Q};\mathcal{D}_{\tau_i}^{S},\bm\theta)$;

Perform meta initialization updates:

\quad\quad$\bm\theta\leftarrow\bm\theta-\frac{\gamma_{1}}{K}\sum_{i=1}^{K}\nabla_{\bm\theta}\mathcal{L}(\mathcal{D}_{\tau_i}^{Q};\mathcal{D}_{\tau_i}^{S},\bm\theta_i)$;

Update the iteration number: $t\leftarrow t+1$;

\If{t\%u=0}{

\tcp{\textcolor{blue}{\textbf{\texttt{Follower Gradient Ascent}}}}

Compute the baseline for the follower:

$\mathcal{V}\approx\frac{1}{K}\sum_{k=1}^K\mathcal{L}(D_{\tau_k}^{Q};\bm\theta)$;

Perform task distribution $p_{\bm\phi}(\tau)$ updates:

$\bm\phi\leftarrow\bm\phi+\frac{\gamma_{2}}{K}\sum_{k=1}^{K}
        \Big[[\mathcal{L}(D_{\tau_k}^{Q};\bm\theta)-\mathcal{V}]\nabla_{\bm\phi}\ln p_{\bm\phi}(\tau_k)
        +\lambda\nabla_{\bm\phi}\ln p_{\bm\phi}(\tau_{k}^{-M})\Big]$.

}}
\caption{Adversarially Task Robust CNP}
\label{arcnp_pseudo}
\end{algorithm}

\subsection{Players' Order in Decision-making}

\begin{remark}[Optimization Order and Solutions]\label{aysmetric_game}
With the risk function $\mathcal{J}(\bm\theta,\bm\phi)$ convex \textit{w.r.t.} $\forall\bm\theta\in\bm\Theta$ and concave \textit{w.r.t.} $\forall\bm\phi\in\bm\Phi$, swapping the order of two players $\{\mathcal{P}_1,\mathcal{P}_2\}$ results in the same equilibrium:
$$\min_{\bm\theta\in\bm\Theta}\max_{\bm\phi\in\bm\Phi}\mathcal{J}(\bm\theta,\bm\phi)=\max_{\bm\phi\in\bm\Phi}\min_{\bm\theta\in\bm\Theta}\mathcal{J}(\bm\theta,\bm\phi).$$
When the convex-concave assumption does not hold, the order generally results in completely different solutions:
$$\min_{\bm\theta\in\bm\Theta}\max_{\bm\phi\in\bm\Phi}\mathcal{J}(\bm\theta,\bm\phi)\neq\max_{\bm\phi\in\bm\Phi}\min_{\bm\theta\in\bm\Theta}\mathcal{J}(\bm\theta,\bm\phi).$$
\end{remark}

The \textbf{Remark} \ref{aysmetric_game} indicates the game is asymmetric \textit{w.r.t.} the players' decision-making order.
This work abstracts the decision-making process for the distribution adversary and the meta-learner in a sequential game.
Hence, the order inevitably influences the solution concept.
Actually, we take MAML as an example to implement the adversarially task robust meta-learning framework.
As noted in Algorithm \ref{armaml_pseudo}, updating the distribution adversary's parameter requires the evaluation of task-specific model parameters, which are not available in the initialization.
The traditional game theory no longer applies to our studied case, and simple counter-examples such as rock/paper/scissors show no equilibrium in practice.
Hence, we pre-determine the order of decision-makers for implementation, as reported in the main paper.

\subsection{Benefits of the Explicit Generative Modeling}
The use of flow-based generative models is beneficial, as the density distribution function is accessible in numerical analysis, allowing one to understand the behavior of the distribution adversary. 
Our training style is adversarial, and the equilibrium is approximately obtained in a gradient update way.
As the stationary point works for both the distribution adversary and the meta player, the resulting meta player is the direct solution to robust fast adaptation.
All of these have been validated from learned probability task densities and entropies.
We treat this as a side product of the method.

\subsection{Set-up of Base Distributions}
This work does not create new synthetic tasks and only considers the regression and continuous control cases.
Our suggestion of a comprehensive base distribution is designed to cover a wide range of scenarios, making it a practical solution in most default setups.
Through the optimization of the distribution adversary, the meta-learner converges to a distribution that focuses on more challenging scenarios and lowers the importance of trivial cases.
Even though enlarging the scope of the scenarios can result in additional computational costs, generative modeling of the task distribution captures more realistic feedback from adaptation performance.
In other words, our study is based on the hypothesis that when the task space is vast enough, the subpopulation shift is allowed under a certain constraint.

\subsection{Related Work Summary}
As far as we know, robust fast adaptation is an underexplored research issue in the field.
In principle, we include the latest SOTA \citep{wang2023sim,collins2020task} and typical SOTA \citep{sagawa2019distributionally} methods to compare in this work as illustrated in Table \ref{meta_learning_principles}.
Note that in \citep{wang2023sim}, the idea of increasing the robustness is to introduce the task selector for risky tasks in implementations.
Though there are some curriculum methods \citep{yao2021meta,dennis2020emergent,parker2022evolving} to reschedule the task sampling probability along the learning, our setup focuses on (i) the semantics in the task identifiers to explicitly uncover the task structure from a Stackelberg game and (ii) the robust solution search under a distribution shift constraint.
That is, ours requires capturing the differentiation information from the task identifiers of tasks, such as masses and lengths in the pendulum system identification.
Ours is agnostic to meta-learning methods and avoids task distribution mode collapse, which might happen in adversarial training without constraints.
These configuration and optimization differences drive us to include the risk minimization principle in the comparisons.

\begingroup
\setlength{\tabcolsep}{1.0pt}
\begin{table*}[h!]
  \begin{center}
    \caption{\textbf{A Summary of Used Methods from Diverse Optimization Principles and Corresponding Properties in Meta-Learning.}
    These are implemented within the empirical/expected risk minimization (ERM), the distributionally robust risk minimization (DRM), and the adversarially task robust risk minimization (ARM).
    Here, MAML works as the backbone method to illustrate the difference.
    }
    \label{meta_learning_principles}
    \begin{tabular}{|c|c|c|c|}
      \toprule 
      Method &Task Distribution &Optimization Objective & Adaptation Robustness \\
      \toprule 
      MAML \citep{finn2017model} & Fixed  & $\min_{\bm\theta\in\bm\Theta}\mathbb{E}_{\tau\sim p(\tau)}\Big[\mathcal{L}(\mathcal{D}_{\tau}^{Q},\mathcal{D}_{\tau}^{S};\bm\theta)\Big]$ & $--$ \\
      \midrule 
      DR-MAML \citep{wang2023sim} & Fixed  & $\min_{\bm\theta\in\bm\Theta}\mathbb{E}_{p_{\alpha}(\tau;\bm\theta)}
        \Big[\mathcal{L}(\mathcal{D}_{\tau}^{Q},\mathcal{D}_{\tau}^{S};\bm\theta)
        \Big]$ & Tail Risk \\
      \midrule 
      TR-MAML \citep{collins2020task} & Fixed  & $\min_{\bm\theta\in\bm\Theta}\max_{\tau\in\mathcal{T}}\mathcal{L}(\mathcal{D}_{\tau}^{Q},\mathcal{D}_{\tau}^{S};\bm\theta)$ & Worst Risk \\
      \midrule 
      DRO-MAML \citep{sagawa2019distributionally} & Fixed  & $\min_{\bm\theta\in\bm\Theta}\max_{g\in\mathcal{G}}\mathbb{E}_{p_{g}(\tau)}
        \Big[\mathcal{L}(\mathcal{D}_{\tau}^{Q},\mathcal{D}_{\tau}^{S};\bm\theta)
        \Big]$ & Uncertainty Set \\
      \bottomrule
      AR-MAML (Ours) & Explicit Generation & $\min_{\bm\theta\in\bm\Theta}\max_{\bm\phi\in\bm\Phi}\mathbb{E}_{p_{\bm\phi}(\tau)}\Big[\mathcal{L}(\mathcal{D}_{\tau}^{Q},\mathcal{D}_{\tau}^{S};\bm\theta)\Big]+\lambda\mathbb{E}_{p_{0}(\tau)}\Big[\ln p_{\bm\phi}(\tau)\Big]$ & Constrained Shift \\
      \bottomrule 
    \end{tabular}
  \end{center}
\end{table*}
\endgroup

\section{Robust Fast Adaptation Strategies in the Task Space}
In this section, we recap recent typical strategies for overcoming the negative effect of task distribution shifts.

\textbf{Group Distributionally Robust Optimization (GDRO) for Meta Learning.}
This method \citep{sagawa2019distributionally} considers combating the distribution shifts via the group of worst-case optimization.
In detail, the general meta training objective can be written as follows:
\begin{equation}\label{append:gdro}
    \begin{split}
        \min_{\bm\theta\in\bm\Theta}\Big\{\mathcal{R}(\bm\theta):=\sup_{g\in\mathcal{G}}\mathbb{E}_{p_{g}(\tau)}
        \Big[\mathcal{L}(\mathcal{D}_{\tau}^{Q},\mathcal{D}_{\tau}^{S};\bm\theta)
        \Big]\Big\},
    \end{split}
\end{equation}
where $\mathcal{G}$ denotes a collection of uncertainty sets at a group level.
When tasks of interest are finite, $g$ induces a probability measure $p_{g}(\tau)$ over a subset, which we call the group in the background of meta-learning scenarios.
The optimization step inside the bracket of Eq. (\ref{append:gdro}) describes the selection of the worst group in fast adaptation performance.
By minimizing the worst group risk, the fast adaptation robustness can be enhanced when confronted with the task distribution shift.

\textbf{Tail Risk Minimization ($\text{CVaR}_{\alpha}$) for Meta Learning.}
This method \citep{wang2023sim} is from the risk-averse perspective, and the tail risk measured $\text{CVaR}_{\alpha}$ by is incorporated in stochastic programming.
The induced meta training objective is written as follows:
\begin{equation}
    \begin{split}
        \min_{\bm\theta\in\bm\Theta}\mathbb{E}_{p_{\alpha}(\tau;\bm\theta)}
        \Big[\mathcal{L}(\mathcal{D}_{\tau}^{Q},\mathcal{D}_{\tau}^{S};\bm\theta)
        \Big],
    \end{split}
\end{equation}
where $p_{\alpha}(\tau;\bm\theta)$ represents the normalized distribution over the tail task in fast adaptation performance, which implicitly relies on the meta learner $\bm\theta$.

Also note that in \citep{wang2023sim}, the above equation can be equivalently expressed as an importance weighted one:
\begin{equation}\label{append:cvar_iw}
    \begin{split}
        \min_{\bm\theta\in\bm\Theta}\mathbb{E}_{p(\tau)}
        \Big[\frac{p_{\alpha}(\tau;\bm\theta)}{p(\tau)}\mathcal{L}(\mathcal{D}_{\tau}^{Q},\mathcal{D}_{\tau}^{S};\bm\theta)
        \Big]
        \approx
        \frac{1}{\mathcal{B}}\sum_{b=1}^{\mathcal{B}}\omega(\tau_b;\bm\theta)\mathcal{L}(D_{\tau_b}^{Q},D_{\tau_b}^{S};\bm\theta),
    \end{split}
\end{equation}
where $\omega(\tau_b;\bm\theta)$ is the ratio of the normalized tail task probability and the original task probability.
In this way, Eq. (\ref{append:cvar_iw}) can be connected to the group distributionally robust optimization method, as the worst group is estimated from the tail risk and reflected in the importance ratio.

\section{Stochastic Gradient Estimates}\label{append:sec_sto_est}
Unlike most adversarial generative models whose likelihood is intractable, our design enables the exact computation of task likelihoods, and the stochastic gradient estimate can be achieved as follows:
\begin{equation}
    \begin{split}
        \nabla_{\bm\phi}\mathcal{J}(\bm\theta,\bm\phi)
        =\nabla_{\bm\phi}\int p_{\bm\phi}(\tau)\mathcal{L}(\mathcal{D}_{\tau}^{Q},\mathcal{D}_{\tau}^{S};\bm\theta)d\tau
        +\lambda\nabla_{\bm\phi}\int p_{0}(\tau)\ln p_{\bm\phi}(\tau)d\tau\\
        =\int p_{\bm\phi}(\tau)\left[
        \nabla_{\bm\phi}\ln p_{\bm\phi}(\tau)\mathcal{L}(\mathcal{D}_{\tau}^{Q},\mathcal{D}_{\tau}^{S};\bm\theta)\right]d\tau
        +\lambda\nabla_{\bm\phi}\int p_{0}(\tau)\nabla_{\bm\phi}\ln p_{\bm\phi}(\tau)d\tau\\
        \approx\frac{1}{K}\sum_{k=1}^{K}\Big[
        \mathcal{L}(D_{\tau_k}^{Q},D_{\tau_k}^{S};\bm\theta)\nabla_{\bm\phi}\ln p_{\bm\phi}(\tau_k)+\lambda\nabla_{\bm\phi}\ln p_{\bm\phi}(\tau_{k}^{-M})\Big]
    \end{split}
    \label{append: meta_adversary_obj}
\end{equation}

With the previously mentioned invertible mappings $\mathcal{G}=\{g_i\}_{i=1}^{M}$, where $g_i:\mathcal{T}\to\mathcal{T}\subseteq\mathbb{R}^d$, we describe the transformation of tasks as $\tau^{-M}\mapsto\dots\mapsto\tau^{-1}\mapsto\tau$.
This indicates that the sampled task particle $\tau^{-M}_{k}$ satisfies the equation:
\begin{equation}
    \begin{split}
        \tau_k=g_{M}\circ\dots g_{2}\circ g_{1}(\tau_{k}^{-M})=\texttt{NN}_{\bm\phi}(\tau_{k}^{-M}),
        \quad
        \text{with}
        \
        \tau_k\sim p_{\bm\phi}(\tau),
        \quad
        k\in\{1,\dots,K\}.
    \end{split}
    \label{normalizing_flows_eq}
\end{equation}

\section{Theoretical Understanding of Generating Task Distribution}
Unlike previous curriculum learning or adversarial training in the task space, this work places a distribution shift constraint over the task space.
Also, our setup tends to create a subpopulation shift under a trust region.

\textbf{Subpopulation Shift Constraint as the Regularization.}
The generated task distribution corresponds to the best response in deteriorating the meta-learner's performance under a tolerant level of the task distribution shift.
Note that the proposed optimization objective $\mathcal{J}(\bm\theta,\bm\phi)$ includes a KL divergence term \textit{w.r.t.} the generated task distribution.

\begin{equation}\label{append: game_meta_obj_kld_cons}
\begin{split}
        \min_{\bm\theta\in\bm\Theta}\max_{\bm\phi\in\bm\Phi}\mathcal{J}(\bm\theta,\bm\phi):=\mathbb{E}_{p_{\bm\phi}(\tau)}\Big[\mathcal{L}(\mathcal{D}_{\tau}^{Q},\mathcal{D}_{\tau}^{S};\bm\theta)\Big]\\
        \textit{s.t.}
        \
        D_{KL}\Big[p_{0}(\tau)\parallel p_{\bm\phi}(\tau)\Big]\leq\delta
\end{split}
\end{equation}

\begin{figure*}[h]
\begin{center}
\centerline{\includegraphics[width=0.9\textwidth]{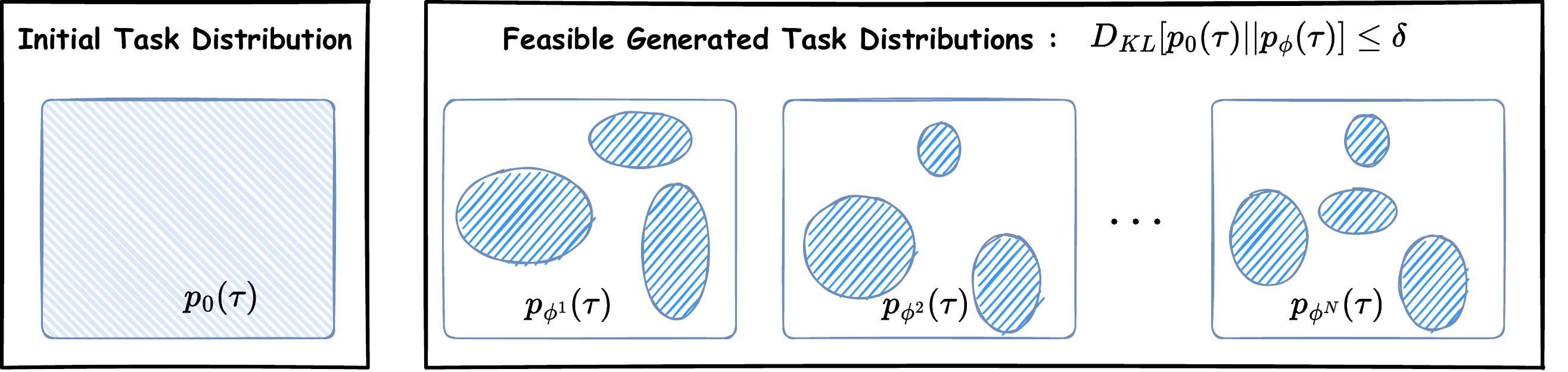}}
\vspace{15pt}
\caption{\textbf{Intuition of distribution shifts and Optimization Steps.}
Left is the initial task distribution, while the generated task distributions within the constraint of the distribution shift are illustrated on the right.
}
\label{append: generated_dist_shift}
\end{center}
\end{figure*}

\textbf{Tolerant Region in distribution shifts.}
Next, we interpret the above-induced optimization objective.
As exhibited in Eq. (\ref{append: game_meta_obj_kld_cons}), the condition constrains the generative task distribution $p_{\bm\phi}(\tau)$ within a neighborhood of the initial task distribution $p_{0}(\tau)$.
This can be viewed as the \textit{tolerant region of task distribution shifts}, which means larger $\delta$ allows more severe distribution shifts concerning the initial task distribution.
As for the role of the distribution adversary, it attempts to seek the task distribution that can handle the strongest task distribution shift under the allowed region, as displayed in Figure \ref{append: generated_dist_shift}.

As mentioned in the main paper, the equivalent unconstrained version can be
\begin{equation}\label{append:uncons_obj}
    \begin{split}
        \min_{\bm\theta\in\bm\Theta}\max_{\bm\phi\in\bm\Phi}\mathcal{J}(\bm\theta,\bm\phi):=\underbrace{\mathbb{E}_{p_{\bm\phi}(\tau)}\Big[\mathcal{L}(\mathcal{D}_{\tau}^{Q},\mathcal{D}_{\tau}^{S};\bm\theta)\Big]}_{\textbf{Adversarial Meta Learning}}
        +\lambda\underbrace{\mathbb{E}_{p_{0}(\tau)}\Big[\ln p_{\bm\phi}(\tau)\Big]}_{\textbf{Distribution Cloning}},
    \end{split}
\end{equation}
where the second penalty is to prevent the generated task distribution from uncontrollably diverging from the initial one.
Hence, it encourages the exploration of crucial tasks in the broader scope while avoiding mode collapse from adversarial training.
The larger the Lagrange multiplier $\lambda$, the weaker the distribution shift in a generation.
In this work, we expect that the tolerant distribution shift can cover a larger scope of shifted task distributions. Hence, $\lambda$ is configured to be a small value as default.

\textbf{Best Response in Constrained Optimization.}
We can also explain the optimization steps in the presence of the Stackelberg game.
Here, we can equivalently rewrite the entangled steps to solve the game as follows:
\begin{subequations}\label{append: game_meta_obj_br}
    \begin{align}
        \min_{\bm\theta\in\bm\Theta}\mathcal{J}(\bm\theta,\bm\phi_{*}(\bm\theta))
        \quad\text{s.t.}
        \
        \bm\phi_{*}(\bm\theta)=\arg\max_{\bm\phi\in\bm\Phi}\mathcal{J}(\bm\theta,\bm\phi)
        \
        \text{and}
        \
        \mathcal{D}_{KL}\Big[p_{0}(\tau)||p_{\bm\phi}(\tau)\Big]\leq\delta
        \\
        \max_{\bm\phi\in\bm\Phi}\mathcal{J}(\bm\theta,\bm\phi)
        \
        \text{and}
        \
        \mathcal{D}_{KL}\Big[p_{0}(\tau)||p_{\bm\phi}(\tau)\Big]\leq\delta,
    \end{align}
\end{subequations}
where Eq. (\ref{append: game_meta_obj_br}.a) indicates the decision-making of the meta learner, while Eq. (\ref{append: game_meta_obj_br}.b) characterizes the decision-making of the distribution adversary.
Particularly, Eq. (\ref{append: game_meta_obj_br}.b) is a constrained sub-optimization problem.

\section{Equilibrium \& Convergence Guarantee}\label{append_sec:convergence}

\begin{figure*}[h]
\begin{center}
\centerline{\includegraphics[width=0.95\textwidth]{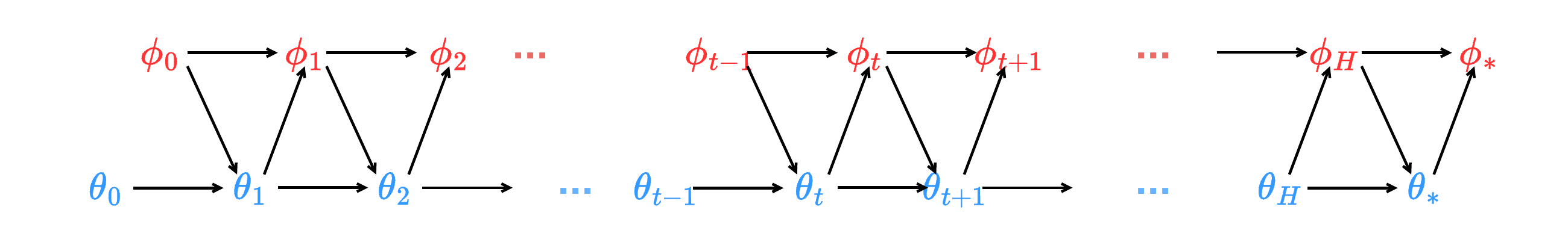}}
\caption{\textbf{Diagram of Stochastic Alternating Gradient Descent Ascent.}
In solving adversarially task robust meta-learning, the updates of separate players' parameters are performed repetitively and can be viewed as a bi-level optimization.  
}
\label{append:fig_alt_sgda}
\end{center}
\end{figure*}

\subsection{Existence of Stackelberg Equilibrium}
\begin{definition}[Global Minimax Point]\label{global_se}
Once the constructed meta-learning Stackelberg game in Eq. (\ref{game_meta_obj}) is solved with the optimal solution as the Stackelberg equilibrium $\{\bm\theta_*,\bm\phi_*\}$, we can naturally obtain the optimal expected risk value $\mathcal{R}_{*}$ and the inequalities $\forall (\bm\theta,\bm\phi)\in\bm\Theta\times\bm\Phi$ as follows:
\begin{equation}
    \begin{split}
        \mathcal{J}(\bm\theta_*,\bm\phi)\leq\mathcal{J}(\bm\theta_*,\bm\phi_*)\leq
        \max_{\hat{\bm\phi}\in\bm\Phi}\mathcal{J}(\bm\theta,\hat{\bm\phi})
        \\
        \mathcal{R}_{*}=\mathcal{J}(\bm\theta_*,\bm\phi_*)
        =\mathbb{E}_{p_{\bm\phi_*}(\tau)}\Big[\mathcal{L}(\mathcal{D}_{\tau}^{Q},\mathcal{D}_{\tau}^{S};\bm\theta_*)\Big]
        +\lambda\mathbb{E}_{p_{0}(\tau)}\Big[\ln p_{\bm\phi_*}(\tau)\Big].
    \end{split}
\end{equation}    
\end{definition}

Note that for the nonconvex-nonconcave min-max optimization problem, there is no general guarantee for the existence of saddle points.
The solution concept in the \textbf{Definition} \ref{global_se} requires no convexity \textit{w.r.t.} the optimization objective and provides a weaker equilibrium, also referred to as the global minimax point \citep{jin2020local}. 

With the \textbf{Assumption} \ref{assum_lipschitz} and the \textbf{Definition} \ref{global_se}, the global Stackelberg equilibrium always exists for the proposed min-max optimization $\min_{\bm\theta\in\bm\Theta}\max_{\bm\phi\in\bm\Phi}\mathcal{J}(\bm\theta,\bm\phi)$.
However, its exact search is NP-hard, and the stochastic optimization practically leads to the local Stackelberg equilibrium in the \textbf{Definition} \ref{local_se}.

\subsection{Convergence Guarantee}

\textbf{Assumption 1 (Lipschitz Smoothness and Compactness)}
\textit{The adversarially task robust meta-learning optimization objective $\mathcal{J}(\bm\theta,\bm\phi)$ is assumed to satisfy}
\begin{enumerate}
    \item 
    \textit{$\mathcal{J}(\bm\theta,\bm\phi)$ with $\forall[\bm\theta,\bm\phi]\in\bm\Theta\times\bm\Phi$ belongs to the class of twice differentiable functions $\mathbb{C}^2$.}
    \item 
    \textit{The norm of block terms inside Hessian matrices $\nabla^2\mathcal{J}(\bm\theta,\bm\phi)$ is bounded, meaning that $\forall[\bm\theta,\bm\phi]\in\bm\Theta\times\bm\Phi$:
    $$\sup\{||\nabla^2_{\bm\theta,\bm\theta}\mathcal{J}||,||\nabla^2_{\bm\theta,\bm\phi}\mathcal{J}||,||\nabla^2_{\bm\phi,\bm\phi}\mathcal{J}||\}\leq L_{\text{max}}.$$}
    \item 
    \textit{The parameter spaces $\bm\Theta\subseteq\mathbb{R}^{d_1}$ and $\bm\Phi\subseteq\mathbb{R}^{d_2}$ are compact with $d_1$ and $d_2$ respectively dimensions of model parameters for two players.}
\end{enumerate}

\textbf{Theorem 1 (Convergence Guarantee)}
\textit{Suppose the \textbf{Assumption} \ref{assum_lipschitz} and the condition of the (local) Stackelberg equilibrium $\Delta(\mathbf{A},\mathbf{B},\mathbf{C},\gamma_1,\gamma_2):=\max\Big\{(1-\gamma_{1}\sigma_{\text{min}})^{2}(1+\gamma_{2}^{2}L_{\text{max}}^{2}),\Big||\gamma_1^2-2\gamma_1\gamma_2+\gamma_1^2\gamma_2^2L_{\text{max}}^2|L_{\text{max}}^2+(1+\gamma_2 L_{\text{max}})^2-2\gamma_1\gamma_2^2\sigma_{\text{min}}(\mathbf{B}^{T}\mathbf{B}\mathbf{C})\Big|\Big\}<\frac{1}{2}$ are satisfied, where $\sigma_{\text{min}}(\cdot)$ denotes the smallest eigenvalues of the corresponding matrix. 
Then we can have the following statements:}
\begin{enumerate}
    \item 
        \textit{The resulting iterated parameters $\{\cdots
        \mapsto
        [\bm\theta_t,\bm\phi_t]^T
        \mapsto
        [\bm\theta_{t+1},\bm\phi_{t+1}]^T
        \mapsto\cdots\}$ are Cauchy sequences.}
    \item
        \textit{The optimization can guarantee at least the linear convergence to the local Stackelberg equilibrium with the rate $\sqrt{\Delta}$.}
\end{enumerate}

\textit{\textbf{Proof of Theorem 1:}}

Remember the learning dynamics as follows, 
\begin{subequations}
    \begin{align}
        \bm\theta_{t+1}\leftarrow\bm\theta_{t}-\gamma_{1}
        \nabla_{\bm\theta}\mathcal{J}(\bm\theta_{t},\bm\phi_{t})
        \\
        \bm\phi_{t+1}\leftarrow\bm\phi_{t}+\gamma_{2}
        \nabla_{\bm\phi}\mathcal{J}(\bm\theta_{t+1},\bm\phi_{t}).
    \end{align}
    \label{append:stochastic_grad_opt}
\end{subequations}
when using the alternating GDA for solving the adaptively robust meta-learning.
Figure \ref{append:fig_alt_sgda} illustrates the iteration rules and steps.

Let $[\bm\theta_*,\bm\phi_*]^T$ be the obtained (local) Stackelberg equilibrium, we denote the difference between the updated model parameter and the equilibrium by $[\hat{\bm\theta}_{t},\hat{\bm\phi}_{t}]^T=[\bm\theta_{t}-\bm\theta_{*},\bm\phi_{t}-\bm\phi_{*}]^T$.
As the utility function $\mathcal{J}(\bm\theta,\bm\phi)$ is Lipschitz smooth, we can perform linearization of $\nabla_{\bm\theta}\mathcal{J}(\bm\theta,\bm\phi)$ and $\nabla_{\bm\phi}\mathcal{J}(\bm\theta,\bm\phi)$ round the resulting stationary point $[\bm\theta_*,\bm\phi_*]^T$ respectively as follows.
\begin{subequations}\label{append:taylor}
    \begin{align}
        \nabla_{\bm\theta}\mathcal{J}(\bm\theta_{t},\bm\phi_{t})
        \approx\nabla_{\bm\theta\bm\theta}^{2}\mathcal{J}(\bm\theta_*,\bm\phi_*)(\bm\theta_t-\bm\theta_*)
        +\nabla_{\bm\theta\bm\phi}^{2}\mathcal{J}(\bm\theta_*,\bm\phi_*)(\bm\phi_t-\bm\phi_*)
        \\
        \nabla_{\bm\phi}\mathcal{J}(\bm\theta_{t+1},\bm\phi_{t})
        \approx\nabla_{\bm\phi\bm\theta}^{2}\mathcal{J}(\bm\theta_*,\bm\phi_*)(\bm\theta_{t+1}-\bm\theta_*)
        +\nabla_{\bm\phi\bm\phi}\mathcal{J}(\bm\theta_*,\bm\phi_*)(\bm\phi_t-\bm\phi_*).
    \end{align}
\end{subequations}
Note that $\nabla_{\bm\phi}\mathcal{J}(\bm\theta_{t+1},\bm\phi_{t})$ in Eq. (\ref{append:taylor}) can be further expressed as
\begin{equation}
\begin{split}
    \nabla_{\bm\phi}\mathcal{J}(\bm\theta_{t+1},\bm\phi_{t})
    \approx
    \nabla_{\bm\phi\bm\theta}^{2}\mathcal{J}(\bm\theta_*,\bm\phi_*)\Big[\bm\theta_t-\bm\theta_*-\gamma_{1}\nabla_{\bm\theta\bm\theta}^2\mathcal{J}(\bm\theta_*,\bm\phi_*)(\bm\theta_t-\bm\theta_*)
    -\gamma_{1}\nabla_{\bm\theta\bm\phi}^{2}\mathcal{J}(\bm\theta_*,\bm\phi_*)(\bm\phi_t-\bm\phi_*)\Big]
    \\
    +\nabla_{\bm\phi\bm\phi}^{2}\mathcal{J}(\bm\theta_*,\bm\phi_*)(\bm\phi_t-\bm\phi_*).    
\end{split}
\end{equation}
with the help of Eq. (\ref{append:stochastic_grad_opt}).
Same as that in the main paper, we write the block terms inside the Hessian matrix $\mathbf{H}_{*}:=\nabla^2\mathcal{J}(\bm\theta_*,\bm\phi_*)$ around $[\bm\theta_*,\bm\phi_*]^T$ as $\begin{bmatrix}
\nabla^2_{\bm\theta\bm\theta}\mathcal{J} &\nabla^2_{\bm\theta\bm\phi}\mathcal{J} \\
\nabla^2_{\bm\phi\bm\theta}\mathcal{J} &\nabla^2_{\bm\phi\bm\phi}\mathcal{J} 
\end{bmatrix}\Big|_{[\bm\theta_*,\bm\phi_*]^T}
:=
\begin{bmatrix}
\mathbf{A} &\mathbf{B} \\
\mathbf{B}^T &\mathbf{C} 
\end{bmatrix}$.

Then we can naturally derive the new form of learning dynamics as follows:
\begin{subequations}\label{append:converted_dynamics}
    \begin{align}
        \hat{\bm\theta}_{t+1}=\hat{\bm\theta}_{t}-\gamma_{1}\mathbf{A}\hat{\bm\theta}_{t}-\gamma_{1}\mathbf{B}\hat{\bm\phi}_{t}
        \\
        \hat{\bm\phi}_{t+1}=\hat{\bm\phi}_{t}+\gamma_2\mathbf{B}^{T}\hat{\bm\theta}_{t}
        -\gamma_1\gamma_2\mathbf{B}^{T}\mathbf{A}\hat{\bm\theta}_{t}
        -\gamma_1\gamma_2\mathbf{B}^{T}\mathbf{B}\hat{\bm\phi}_{t}
        +\gamma_2\mathbf{C}\hat{\bm\phi}_{t}.
    \end{align}
\end{subequations}

Further, for the sake of simplicity, we rewrite the above mentioned matrices as $\mathbf{P}=\mathbf{I}-\gamma_1\mathbf{A}$ and $\mathbf{Q}=\mathbf{I}+\gamma_2\mathbf{C}$.
Equivalently, Eq. (\ref{append:converted_dynamics}) can be written in the matrix form:
\begin{equation}\label{append:mat_converted_dynamics}
    \begin{split}
        \begin{bmatrix}
            \hat{\bm\theta}_{t+1}
            \\
            \hat{\bm\phi}_{t+1}
        \end{bmatrix}
        =\begin{bmatrix}
            \mathbf{P} & -\gamma_1\mathbf{B}\\
            \gamma_2\mathbf{B}^{T}-\gamma_1\gamma_2\mathbf{B}^{T}\mathbf{A} &\mathbf{Q}-\gamma_1\gamma_2\mathbf{B}^{T}\mathbf{B}
        \end{bmatrix}
        \cdot
        \begin{bmatrix}
            \hat{\bm\theta}_{t}
            \\
            \hat{\bm\phi}_{t}
        \end{bmatrix}.
    \end{split}
\end{equation}

Furthermore, we consider the expression of the updated parameters' norm:
\begin{subequations}
    \begin{align}
        ||\hat{\bm\theta}_{t+1}||_{2}^{2}+||\hat{\bm\phi}_{t+1}||_{2}^{2}
        =||\mathbf{P}\hat{\bm\theta}_{t}-\gamma_1\mathbf{B}\hat{\bm\phi}_{t}||_{2}^{2}
        +||(\gamma_2\mathbf{B}^{T}-\gamma_1\gamma_2\mathbf{B}^{T}\mathbf{A})\hat{\bm\theta}_{t}
        +(\mathbf{Q}-\gamma_1\gamma_2\mathbf{B}^{T}\mathbf{B})\hat{\bm\phi}_{t}||_{2}^{2}\\
         =||\mathbf{P}\hat{\bm\theta}_{t}-\gamma_1\mathbf{B}\hat{\bm\phi}_{t}||_{2}^{2}
        +||\gamma_2\mathbf{B}^{T}\mathbf{P}\hat{\bm\theta}_{t}
        +(\mathbf{Q}-\gamma_1\gamma_2\mathbf{B}^{T}\mathbf{B})\hat{\bm\phi}_{t}||_{2}^{2}\\
        \leq
        2\Big[||\mathbf{P}\hat{\bm\theta}_{t}||_{2}^{2}+\gamma_{1}^{2}||\mathbf{B}\hat{\bm\phi}_{t}||_{2}^{2}\Big]
        +2\Big[\gamma_2^2||\mathbf{B}^{T}\mathbf{P}\hat{\bm\theta}_{t}||_2^2
        +||(\mathbf{Q}-\gamma_1\gamma_2\mathbf{B}^{T}\mathbf{B})\hat{\bm\phi}_{t}||_2^2\Big]\\
        =2\Big[||\mathbf{P}\hat{\bm\theta}_{t}||_{2}^{2}+\gamma_2^2||\mathbf{B}^{T}\mathbf{P}\hat{\bm\theta}_{t}||_2^2\Big]+2\Big[\gamma_{1}^{2}||\mathbf{B}\hat{\bm\phi}_{t}||_{2}^{2}+||(\mathbf{Q}-\gamma_1\gamma_2\mathbf{B}^{T}\mathbf{B})\hat{\bm\phi}_{t}||_2^2\Big],
    \end{align}
\end{subequations}
where the last inequality makes use of the Cauchy–Schwarz inequality trick $||\mathbf{a}+\mathbf{b}||_2^2\leq 2(||\mathbf{a}||_2^2+||\mathbf{b}||_2^2)$, $\forall \mathbf{a},\mathbf{b}\in\mathbb{R}^d$.

For terms concerning $\bm\theta$, we can have the following evidence:
\begin{subequations}\label{append:theta_eq}
    \begin{align}
        ||\mathbf{P}\hat{\bm\theta}_{t}||_{2}^{2}+\gamma_2^2||\mathbf{B}^{T}\mathbf{P}\hat{\bm\theta}_{t}||_2^2
        =\hat{\bm\theta}_{t}^{T}\Big(\mathbf{P}^{T}\mathbf{P}+\gamma_2^2\mathbf{P}^{T}\mathbf{B}\mathbf{B}^{T}\mathbf{P}\Big)\hat{\bm\theta}_{t}\\
        \leq(1+\gamma_2^2L_{\text{max}}^2)\hat{\bm\theta}_{t}^{T}\mathbf{P}^{T}\mathbf{P}\hat{\bm\theta}_{t}\\
        \leq(1+\gamma_2^2L_{\text{max}}^2)(1-\gamma_1\sigma_{\text{min}}(\mathbf{A}))||\hat{\bm\theta}_{t}||_2^2,
    \end{align}
\end{subequations}
where the last two inequalities make use of the assumptions and tricks that (i) the boundness assumption $\mathbf{B}\leq L_{\text{max}}$ and (ii) the trait of the symmetric matrix $||\mathbf{P}||_2=||\mathbf{I}-\gamma_1\mathbf{A}||_2\leq 1-\gamma_1\sigma_{\text{min}}(\mathbf{A})$.

For terms concerning $\bm\phi$, we can have the following evidence:
\begin{subequations}\label{append:phi_eq}
    \begin{align}
        \gamma_{1}^{2}||\mathbf{B}\hat{\bm\phi}_{t}||_{2}^{2}+||(\mathbf{Q}-\gamma_1\gamma_2\mathbf{B}^{T}\mathbf{B})\hat{\bm\phi}_{t}||_2^2
        =\hat{\bm\phi}_{t}^{T}\Big(\gamma_{1}^{2}\mathbf{B}^{T}\mathbf{B}+(\mathbf{Q}-\gamma_1\gamma_2\mathbf{B}^{T}\mathbf{B})^{T}(\mathbf{Q}-\gamma_1\gamma_2\mathbf{B}^{T}\mathbf{B})\Big)\hat{\bm\phi}_{t}
        \\
        =\hat{\bm\phi}_{t}^{T}\mathbf{B}^{T}\Big(\gamma_1^2\mathbf{I}+\gamma_1^2\gamma_2^2\mathbf{B}\mathbf{B}^{T}\Big)\mathbf{B}\hat{\bm\phi}_{t}
        +\hat{\bm\phi}_{t}^{T}\mathbf{Q}^{T}\mathbf{Q}\hat{\bm\phi}_{t}
        -2\gamma_1\gamma_2\hat{\bm\phi}_{t}^{T}(\mathbf{B}^{T}\mathbf{B}\mathbf{Q})\hat{\bm\phi}_{t}
        \\
        =\hat{\bm\phi}_{t}^{T}\mathbf{B}^{T}\Big(\gamma_1^2\mathbf{I}+\gamma_1^2\gamma_2^2\mathbf{B}\mathbf{B}^{T}\Big)\mathbf{B}\hat{\bm\phi}_{t}
        +\hat{\bm\phi}_{t}^{T}\mathbf{Q}^{T}\mathbf{Q}\hat{\bm\phi}_{t}
        -2\gamma_1\gamma_2\hat{\bm\phi}_{t}^{T}\mathbf{B}^{T}\mathbf{B}\hat{\bm\phi}_{t}
        -2\gamma_2\gamma_2^2\hat{\bm\phi}_{t}^{T}\Big(\mathbf{B}^{T}\mathbf{B}\mathbf{C}\Big)\hat{\bm\phi}_{t}
        \\
        =\hat{\bm\phi}_{t}^{T}\mathbf{B}^{T}\Big((\gamma_1^2-2\gamma_1\gamma_2)\mathbf{I})+\gamma_1^2\gamma_2^2\mathbf{B}\mathbf{B}^{T}\Big)\mathbf{B}\hat{\bm\phi}_{t}
        +\hat{\bm\phi}_{t}^{T}\mathbf{Q}^{T}\mathbf{Q}\hat{\bm\phi}_{t}
        -2\gamma_2\gamma_2^2\hat{\bm\phi}_{t}^{T}\Big(\mathbf{B}^{T}\mathbf{B}\mathbf{C}\Big)\hat{\bm\phi}_{t}
        \\
        \leq
        \Big||\gamma_1^2-2\gamma_1\gamma_2+\gamma_1^2\gamma_2^2L_{\text{max}}^2|L_{\text{max}}^2+(1+\gamma_2 L_{\text{max}})^2-2\gamma_1\gamma_2^2\sigma_{\text{min}}(\mathbf{B}^{T}\mathbf{B}\mathbf{C})\Big|||\hat{\bm\phi}_{t}||_2^2.
    \end{align}
\end{subequations}

\subsection{Convergence Properties}

\textit{\textbf{Property (1):}}
Given the assumption $\Delta(\mathbf{A},\mathbf{B},\mathbf{C},\gamma_1,\gamma_2):=\max\Big\{(1-\gamma_{1}\sigma_{\text{min}})^{2}(1+\gamma_{2}^{2}L_{\text{max}}^{2}),\Big||\gamma_1^2-2\gamma_1\gamma_2+\gamma_1^2\gamma_2^2L_{\text{max}}^2|L_{\text{max}}^2+(1+\gamma_2 L_{\text{max}})^2-2\gamma_1\gamma_2^2\sigma_{\text{min}}(\mathbf{B}^{T}\mathbf{B}\mathbf{C})\Big|\Big\}<\frac{1}{2}$, we can draw up the deduction that $||\hat{\bm\theta}_{t+1}||_{2}^{2}+||\hat{\bm\phi}_{t+1}||_{2}^{2}\leq\Delta\Big(\hat{\bm\theta}_{t}||_{2}^{2}+||\hat{\bm\phi}_{t}\Big)$ with $\Delta<1$ directly from Eq.s (\ref{append:theta_eq}) and (\ref{append:phi_eq}).

For ease of simplicity, we rewrite model parameters as $\mathbf{z}=[\bm\theta,\bm\phi]^T\in\mathcal{Z}=\bm\Theta\times\bm\Phi$ and $\hat{\mathbf{z}}=[\hat{\bm\theta},\hat{\bm\phi}]^T$.
The deduction can be equivalently expressed as $||\hat{\mathbf{z}}_{t+1}||_{2}\leq\sqrt{\Delta}||\hat{\mathbf{z}}_{t}||_2<||\hat{\mathbf{z}}_{t}||_2$.

Then $\forall\epsilon>0$, there always exists an integer $N\in\mathbb{N}^+$ such that for all $m>n>N$,
\begin{subequations}
    \begin{align}
        ||\mathbf{z}_m-\mathbf{z}_n||_2
        =||\hat{\mathbf{z}}_m-\hat{\mathbf{z}}_n||_2
        \leq\sum_{t=1}^{m-n}||\hat{\mathbf{z}}_{n+t}-\hat{\mathbf{z}}_{n+t-1}||_2
        \leq\Big[\sum_{t=1}^{m-n}\sqrt{\Delta}^{n+k}\Big]
        \Big(||\hat{\mathbf{z}}_{0}||_2\Big)
        <\epsilon,
    \end{align}
\end{subequations}
where the inequality can be obviously satisfied when $N$ is large enough.
This implies the iteration sequence $\{z_t\}_{t=0}^{H}$ is Cauchy.

\textit{\textbf{Property (2):}}
Given the assumption $\Delta:=\max\Big\{(1-\gamma_{1}\sigma_{\text{min}})^{2}(1+\gamma_{2}^{2}L_{\text{max}}^{2}),\Big||\gamma_1^2-2\gamma_1\gamma_2+\gamma_1^2\gamma_2^2L_{\text{max}}^2|L_{\text{max}}^2+(1+\gamma_2 L_{\text{max}})^2-2\gamma_1\gamma_2^2\sigma_{\text{min}}(\mathbf{B}^{T}\mathbf{B}\mathbf{C})\Big|\Big\}<\frac{1}{2}$ and the property (1), we can derive that $||\hat{\mathbf{z}}_{t}||_2\leq(\sqrt{\Delta})^{t}||\hat{\mathbf{z}}_0||_2$.
After performing the limit operation, we can have:
\begin{equation}
    \lim_{t\to\infty}||\hat{\mathbf{z}}_{t}||_2\leq\lim_{t\to\infty}(\sqrt{\Delta})^{t}||\hat{\mathbf{z}}_0||_2=0
    \implies
    \lim_{t\to\infty} [\bm\theta_t,\bm\phi_t]^T=[\bm\theta_*,\bm\phi_*]^T.
\end{equation}
Hence, the adopted optimization strategy can guarantee at least the linear convergence with the rate $\sqrt{\Delta}$ to the (local) Stackelberg equilibrium.

\section{Generalization Bound}\label{append_sec:generalization}

Note that the task distribution is adaptive and learnable, and we take interest in the generalization in the context of generative task distributions.
It is challenging to perform direct analysis. 
Hence, we propose to exploit the importance of the weighting trick.
To do so, we first recap the reweighted generalization bound from \citep{cortes2010learning} as \textbf{Lemma} \ref{lemma_reweighted_gbound}.
The sketch of proofs mainly consists of the importance-weighted generalization bound and estimates of the importance weights' range.

\subsection{Importance Weighted Generalization Bound}
\begin{lemma}[Generalization Bound of Reweighted Risk \citep{cortes2010learning}]\label{lemma_reweighted_gbound}
    Given a risk function $\mathcal{L}$ and arbitary hypothesis $\bm\theta$ inside the hypothesis space $\bm\theta\in\bm\Theta$ together with the pseudo-dimension $\mathcal{C}=\text{Pdim}(\{\mathcal{L}(\cdot;\bm\theta):\bm\theta\in\bm\Theta\})$ in \citep{pollard1984convergence} and the importance variable $\omega(\tau)$, then the following inequality holds with a probability $1-\delta$ over samples $\{\tau_1,\tau_2,\dots,\tau_K\}$:
    \begin{equation}
        \begin{split}
            R_p^{\omega}(\bm\theta)
            \leq
            \hat{R}_p^{\omega}(\bm\theta)
            +2^{\frac{5}{4}}V_{p,\hat{p}}\left[\omega(\tau)\mathcal{L}(\mathcal{D}_{\tau}^{Q},\mathcal{D}_{\tau}^{S};\bm\theta)\right]\left(\frac{\mathcal{C}\ln\frac{2Ke}{\mathcal{C}}+\ln\frac{4}{\delta}}{K}\right)^{\frac{3}{8}},
        \end{split}
    \end{equation}
where $V_{p,\hat{p}}\left[\omega(\tau)\mathcal{L}(\mathcal{D}_{\tau}^{Q},\mathcal{D}_{\tau}^{S};\bm\theta)\right]=\max\left\{\sqrt{\mathbb{E}_{p}[\omega^2(\tau)\mathcal{L}^{2}(\mathcal{D}_{\tau}^{Q},\mathcal{D}_{\tau}^{S};\bm\theta)]},\sqrt{\mathbb{E}_{\hat{p}}[\omega^2(\tau)\mathcal{L}^{2}(\mathcal{D}_{\tau}^{Q},\mathcal{D}_{\tau}^{S};\bm\theta)]}\right\}$ with $p$ the exact task distribution, $\hat{p}$ the empirical task distribution, and $\mathbb{E}_{p}[\omega(\tau)]=1$.
\end{lemma}

Further we denote the expected risk of interest by $R_p^{\omega}(\bm\theta_*)=\mathbb{E}_{p_{\bm\phi}(\tau)}\left[\mathcal{L}(\mathcal{D}_{\tau}^{Q},\mathcal{D}_{\tau}^{S};\bm\theta_*)\right]=\mathbb{E}_{p_{0}(\tau)}\left[\omega(\tau)\mathcal{L}(\mathcal{D}_{\tau}^{Q},\mathcal{D}_{\tau}^{S};\bm\theta_*)\right]$.
The corresponding empirical risk is $\hat{R}_p^{\omega}(\bm\theta_*):=
        \frac{1}{K}\sum_{k=1}^{K}
        \omega(\tau_k^0)\mathcal{L}(D_{\tau_k^0}^{Q},D_{\tau_k^0}^{S};\bm\theta_*)$.
Details of these terms' relations are attached as below.
\begin{equation}
    \begin{split}
        \mathbb{E}_{p_{0}(\tau)}\left[\frac{p_{\bm\phi}(\tau)}{p_{0}(\tau)}\mathcal{L}(\mathcal{D}_{\tau}^{Q},\mathcal{D}_{\tau}^{S};\bm\theta_*)\right]
        =\mathbb{E}_{p_{0}(\tau)}\left[\omega(\tau)\mathcal{L}(\mathcal{D}_{\tau}^{Q},\mathcal{D}_{\tau}^{S};\bm\theta_*)\right]
        \approx
        \frac{1}{K}\sum_{k=1}^{K}
        \omega(\tau_k^0)\mathcal{L}(D_{\tau_k^0}^{Q},D_{\tau_k^0}^{S};\bm\theta_*),
        \
        \text{with}
        \
        \tau_k^0\sim p_{0}(\tau).
    \end{split}
\end{equation}

\subsection{Formal Adversarial Generalization Bounds}

Let us further denote the generated sequence of task identifiers by $\tau_k^{-M}\mapsto\tau_k^{-M+1}\mapsto\dots\mapsto\tau_k^0$ and consider the uniform distribution as the base distribution cases $p_{0}(\tau_k^{-M})=p_{0}(\tau_k^0)$, then we can have the following equation:
\begin{subequations}
    \begin{align}
    \omega(\tau_k^0)=\exp\left(\ln p_{\bm\phi}(\tau_k^0)-\ln p_{0}(\tau_k^0)\right)
    \\
    =\exp\left(\ln p_{0}(\tau_k^{-M})+\sum_{i=1}^{M}\ln\left|\det\frac{\partial g_{i}^{-1}}{\partial \tau_k^{-M+i}}\right|-\ln p_{0}(\tau_k^0)\right)
    \\
    =\exp\left(\ln p_{0}(\tau_k^0)+\sum_{i=1}^{M}\ln\left|\det\frac{\partial g_{i}^{-1}}{\partial \tau_k^{-M+i}}\right|-\ln p_{0}(\tau_k^0)\right)
    \\
    =\exp\left(\sum_{i=1}^{M}\ln\left|\det\frac{\partial g_{i}^{-1}}{\partial \tau_k^{-M+i}}\right|\right).
    \end{align}
\end{subequations}

Note that the invertible functions inside normalizing flows are assumed to hold the bi-Lipschitz property.
For the estimate of the term $\left|\det\frac{\partial g_{i}^{-1}}{\partial \tau_k^{-M+i}}\right|$, we directly apply Hadamard's inequality \citep{maz1999jacques} to obtain:
\begin{equation}
    \begin{split}
        \left|\det\frac{\partial g_{i}^{-1}}{\partial \tau_k^{-M+i}}\right|
        \leq
        \prod_{s=1}^{d}||\frac{\partial g_{i}^{-1}}{\partial \tau_k^{-M+i}}v_s||
        \leq
        ||\frac{\partial g_{i}^{-1}}{\partial \tau_k^{-M+i}}||^d
        \leq
        \ell_{a}^d,
        \
        \forall m\in\{1,2,\dots,M\}
        \\
        \Longrightarrow
        \ln\left|\det\frac{\partial g_{i}^{-1}}{\partial \tau_k^{-M+i}}\right|
        \leq
        d\ln \ell_{a}
        \\
        \Longrightarrow
        \sum_{i=1}^M\ln\left|\det\frac{\partial g_{i}^{-1}}{\partial \tau_k^{-M+i}}\right|
        \leq
        dM\ln \ell_{a},
    \end{split}
\end{equation}
where $v_s$ is a collection of unit eigenvectors of the considered Jacobian.
The above implies that $\omega(\tau_k^0)\leq \ell_{a}^{Md}$.

Now, we put the above derivations together and present the formal adversarial bound as follows.
\begin{equation}
    \max\left\{\mathbb{E}_{p_0(\tau)}[\omega^2(\tau)\mathcal{L}^{2}(\mathcal{D}_{\tau}^{Q},\mathcal{D}_{\tau}^{S};\bm\theta_*)],\mathbb{E}_{\hat{p}_0(\tau)}[\omega^2(\tau)\mathcal{L}^{2}(\mathcal{D}_{\tau}^{Q},\mathcal{D}_{\tau}^{S};\bm\theta_*)]\right\}
    \leq
    \ell_{a}^{2Md}\sup_{\tau\in\mathcal{T}}|\mathcal{L}(\mathcal{D}_{\tau}^{Q},\mathcal{D}_{\tau}^{S};\bm\theta_*)|^2
\end{equation}

This formulates the generalization bound of meta learners under the learned adversarial task distribution.

\textbf{Theorem 2 (Generalization Bound with the Distribution Adversary)}
\textit{Given the pretrained normalizing flows $\{g_i\}_{i=1}^M$, where $g_i$ is presumed to be $(\ell_{a}, \ell_{b})$-bi-Lipschitz, the pretrained meta learner $\bm\theta_*$ from the hypothesis space $\bm\theta_*\in\bm\Theta$ together with the pseudo-dimension $\mathcal{C}=\text{Pdim}(\{\mathcal{L}(\cdot;\bm\theta):\bm\theta\in\bm\Theta\})$ in \citep{pollard1984convergence}, we can derive the generalization bound when the initial task distribution $p$ is uniform.
    \begin{equation}
        \begin{split}
            R_p^{\omega}(\bm\theta)
            \leq
            \hat{R}_p^{\omega}(\bm\theta)
            +\Upsilon(\mathcal{T})\left(\frac{\mathcal{C}\ln\frac{2Ke}{\mathcal{C}}+\ln\frac{4}{\delta}}{K}\right)^{\frac{3}{8}},
        \end{split}
    \end{equation}
the inequality holds with a probability $1-\delta$ over samples $\{\tau_1,\tau_2,\dots,\tau_K\}$, where the constant means $\Upsilon(\mathcal{T})=2^{\frac{5}{4}}\ell_{a}^{2Md}\sup_{\tau\in\mathcal{T}}|\mathcal{L}(\mathcal{D}_{\tau}^{Q},\mathcal{D}_{\tau}^{S};\bm\theta_*)|^2$.}

\section{Explicit Generative Task Distributions}

Once the meta training process is finished, we can direct access the generative task distributions from the pretrained normalizing flows.
For any task $\tau^0$, we still assume the generated task sequence is $\tau^{-M}\mapsto\tau^{-M+1}\mapsto\dots\mapsto\tau^0$ after the invertible transformations $\{g_i\}_{i=1}^M$.
Then, we can obtain the exact likelihood of sampling the task $\tau$ as:
\begin{equation}
    \begin{split}
        p_{\bm\phi_*}(\tau^0)=\frac{p_{0}(\tau^{-M})}{\prod_{i=1}^{M}\left|\det\frac{\partial g_{i}^{-1}}{\partial \tau^{-M+i}}\right|},
    \end{split}
\end{equation}
where the product of the Jacobian scales the task probability density.

\section{Evolution of Entropies in Task Distributions}
One of the primary benefits of employing normalizing flows in generating task distributions is the tractable likelihood, leaving it possible to analyze the entropy of task distributions.
Note that \textbf{Remark} \ref{remark:entropy} in the main paper characterizes the way of computing the generative task distribution entropy.

\textbf{Remark 1 (Entropy of the Generated Task Distribution)}
\textit{Given the generative task distribution $p_{\bm\phi_*}(\tau)$, we can derive its entropy from the initial task distribution $p_0(\tau)$ and normalizing flows $\mathcal{G}=\{g_i\}_{i=1}^{M}$:}
    \begin{equation}
        \begin{split}
            \mathbb{H}\Big[p_{\bm\phi_*}(\tau)\Big]
            =\mathbb{H}\Big[p_{0}(\tau)\Big]
            +\int p_{0}(\tau)\left[\sum_{i=1}^{M}\ln\left|\det\frac{\partial g_{i}}{\partial \tau^{i}}\right|\right]d\tau.
        \end{split}
    \end{equation}
\textit{The above implies that the generated task distribution entropy is governed by the change of task identifiers in the probability measure of task space.}

Next is to provide the proof \textit{w.r.t.} the above Remark.

\textit{\textbf{Proof:}}
\textit{Given the initial distribution $p_0(\tau)$ and the sampled task $\tau^0$, we know the transformation within the normalizing flows:}
$$\tau^M=g_{M}\circ\dots g_{2}\circ g_{1}(\tau^0)=\texttt{NN}_{\bm\phi}(\tau^0).$$
\textit{Also, the density function after transformations is: }
$$\ln p_{\bm\phi}(\tau^{M})=\ln p_{0}(\tau^{0})
-\sum_{i=1}^{M}\ln\left|\det\frac{\partial g_{i}}{\partial \tau^{i}}\right|.$$
\textit{With the above information, we can naturally derive the following equations:}
\begin{equation}
    \begin{split}
        \mathbb{H}\Big[p_{\bm\phi_*}(\tau^{M})\Big]
        =-\int p_{\bm\phi_*}(\tau^{M})\ln p_{\bm\phi_*}(\tau^{M})d\tau^{M}\\
        =-\int p_{0}(\tau^{0})\left[\ln p_{0}(\tau^{0})
        -\sum_{i=1}^{M}\ln\left|\det\frac{\partial g_{i}}{\partial \tau^{i}}\right|\right]d\tau^{0}\\
        =\mathbb{H}\Big[p_{0}(\tau^{0})\Big]
        +\int p_{0}(\tau^{0})\left[\sum_{i=1}^{M}\ln\left|\det\frac{\partial g_{i}}{\partial \tau^{i}}\right|\right]d\tau^{0}.
    \end{split}
\end{equation}
\textit{This completes the proof of \textbf{Remark} \ref{remark:entropy}.}

\section{Experimental Set-up \& Implementation Details}\label{append:exp setup}

\subsection{Meta Learning Benchmarks}
\begingroup
\setlength{\tabcolsep}{3.0pt}
\begin{table*}[h!]
  \begin{center}
    \caption{\textbf{A Summary of Benchmarks in Evaluation.}
    We report task identifiers in configuring tasks, corresponding initial ranges and types of base distributions.
    $\mathcal{U}$ and $\mathcal{N}$ respectively denote the uniform and the normal distributions, with distribution parameters inside the bracket.
    }
    \label{append:benchmark_info}
    \begin{tabular}{|c|c|c|c|}
      \toprule 
      Benchmarks &Task Identifiers &Identifier Range &Initial Distribution \\
      \toprule 
      Sinusoid-U &amplitude $a$ and phase $b$ &$(a,b)\sim[0.1,5.0]\times[0.0,\pi]$ & $\mathcal{U}([0.1,5.0]\times[0.0,\pi])$\\
      \bottomrule 
       Sinusoid-N &amplitude $a$ and phase $b$ &$(a,b)\sim[0.1,5.0]\times[0.0,\pi]$ &$\mathcal{N}([2.5,1.5],diag(0.8^2,0.5^2))$\\
      \bottomrule 
      Acrobot-U &pendulum masses $(m_1,m_2)$ &$(m_1,m_2)\sim[0.4,1.6]\times[0.4,1.6]$ & $\mathcal{U}([0.4,1.6]\times[0.4,1.6])$\\
      \midrule 
      Acrobot-N &pendulum masses $(m_1,m_2)$ &$(m_1,m_2)\sim[0.4,1.6]\times[0.4,1.6]$ &$\mathcal{N}([1.0,1.0],diag(0.2^2,0.2^2))$\\
      \midrule 
      Pendulum-U &pendulum mass $m$ and length $l$ &$(m,l)\sim[0.4,1.6]\times[0.4,1.6]$ & $\mathcal{U}([0.4,1.6]\times[0.4,1.6])$\\
      \bottomrule
      Pendulum-N &pendulum mass $m$ and length $l$ &$(m,l)\sim[0.4,1.6]\times[0.4,1.6]$ &$\mathcal{N}([1.0,1.0],diag(0.2^2,0.2^2))$\\
      \midrule
      Point Robot &goal location $(x_1,x_2)$ &$(x_1,x_2)\sim[-0.5,0.5]\times[-0.5,0.5]$ & $\mathcal{U}([-0.5,0.5]\times[-0.5,0.5])$\\
      \midrule 
      Pos-Ant & target position $(x_1,x_2)$ & $(x_1,x_2)\sim[-3.0,3.0]\times[-3.0,3.0]$ & $\mathcal{U}([-3.0,3.0]\times[-3.0,3.0])$\\
      \bottomrule 
    \end{tabular}
  \end{center}
\end{table*}
\endgroup

The following details the meta-learning dataset, and we also refer the reader to the \textbf{List (1)} for the preprocessing of data.

\textbf{Sinusoid Regression.}
In sinusoid regression, each task is equivalent to mapping the input to the output of a sine wave.
Here, the task identifiers are the amplitude $a$ and the phase $b$.
Data points in regression are collected in the way: $10$ data points are uniformly sampled from the interval $[-5.0,5.0]$, coupled with the output $y=a\sin(x-b)$. 
These data points are divided into the support dataset ($5$-shot) and the query dataset.
As for the range of task identifiers and the types of initial distributions, please refer to Table \ref{append:benchmark_info}.
In meta-testing phases, we randomly sample $500$ tasks from the initial task distribution and the generated task distribution to evaluate the performance, and this results in Table \ref{test_regression}.

\textbf{System Identification.}
In the Acrobot System, angles and angular velocities characterize the state of an environment as $[\bm\theta_{1},\bm\theta_{1}^{\prime},\bm\theta_2,\bm\theta_{2}^{\prime}]$.
The goal is to identify the system dynamics, namely the transited state, after selecting Torque action from $\{-1,0,+1\}$.
In the Pendulum System, environment information can be found in the OpenAI gym.
In detail, the observation is $(\cos\bm\theta,\sin\bm\theta,\bm\theta^{\prime})$ with $\bm\theta\in[-\pi,\pi]$, and the continuous action range is $[-2.0, 2.0]$.
The torque is executed on the pendulum body, and the goal is to predict the dynamics given the observation and action.
For both dynamical systems, we use a random policy as the controller to interact with the environment to collect transition samples.
For the few-shot purpose, we sample $200$ transitions from each Markov Decision Process as one batch and randomly split them into the support dataset ($10$-shot transitions) and the query dataset.
In meta training, the meta-batch in iteration is $16$, and the maximum iteration number is $500$.
In meta-testing phases, we randomly sample $100$ tasks respectively from the initial and generated task distributions with each task one batch in evaluation, which results in Table \ref{test_regression}.

\textbf{Meta Learning Continuous Control.}
The environment of reinforcement learning is mainly treated as a Markov decision process \citep{mao2024supported,mao2023supported}.
And the meta RL is about the distribution over environments.
We consider the navigation task to examine the performance of methods in reinforcement learning.
The mission is to guide the robot, e.g., the point robot and the Ant from Mujoco, to move towards the target goal step by step.
Hence, the task identifier is the goal location $(x_1,x_2)$.
The agent performs $20$ episode explorations to identify the environment and enable inner policy gradient updates as fast adaptation.
The environment information, such as transitions and rewards, is accessible at (\url{https://github.com/lmzintgraf/cavia/tree/master/rl/envs}).
Particularly, for the task distribution like $\mathcal{U}([-0.5,0.5]\times[-0.5,0.5])$ in the point robot, we set the dark area in Figure \ref{vis_benchmarks}c as the sparse reward region, discounting the step-wise reward by $0.6$ to area $[-0.25,-0.5]\times[0.25,0.5]$ and $0.4$ to aera $[0.25,0.5]\times[-0.25,-0.5]$.
In meta training, the meta-batch in the iteration is 20 for the point robot, and 40 for Ant, and the model is trained for up to 500 meta-iterations.
In meta-testing, we randomly sampled 100 MDPs, specifically from the initial and generated task distributions.
For each MDP, we run 20 episodes as the support dataset and compute the accumulated returns after fast adaptation.

The general implementation details are retained the same as those in MAML (\url{https://github.com/tristandeleu/pytorch-maml-rl}) and CAVIA (\url{https://github.com/lmzintgraf/cavia}).

\subsection{Modules in Pytorch}
To enable the implementation of our developed framework, we report the neural modules implemented in Pytorch.
The whole model consists of the distribution adversary and the meta learner.
For the sake of clarity, we attach the separate modules below.
Our implementation also relies on the normalizing flow package available at (\url{https://github.com/VincentStimper/normalizing-flows/tree/master/normflows}).

\begin{lstlisting}[language=Python, caption=the Distribution Adversary and the Meta Learner.]
"""
Neural network models for the regression experiments with adaptively robust maml.
"""

import math
import torch
import torch.nn.functional as F
from torch import nn
import normflows as nf
from normflows.flows import Planar, Radial

################################################################################################################
    # This part is to introduce the flow module to transform random variables. --> Distribution Adversary
################################################################################################################


class Distribution_Adversary(nn.Module):
    def __init__(self, 
                 q0,  
                 latent_size,
                 num_latent_layers,
                 flow_type,
                 hyper_range, 
                 device
                 ):
        '''
        q0: base distribution of task parameters
        latent_size: the dimension of the latent variable
        num_latent_layers: number of layers in NFs
        flow_type: types of NFs
        hyper_range: range of task hyper-parameters, tensor shape [dim_z, 2] e.g. [[4.9, 0.1], [3.0, -1.0]] -> [param_range, range_min]
        device: 'cuda' or 'cpu'
        '''
        
        super(Distribution_Adversary, self).__init__()
        
        self.q0 = q0 
        self.latent_size = latent_size 
        self.num_latent_layers = num_latent_layers 
        self.flow_type = flow_type 
        self.hyper_range = hyper_range
        self.device = device
        
        if flow_type == 'Planar_Flow':
            flows = [Planar(self.latent_size) for k in range(self.num_latent_layers)]
        elif flow_type == 'Radial_Flow':
            flows = [Radial(self.latent_size) for k in range(self.num_latent_layers)]
            
        self.nfm = nf.NormalizingFlow(q0=self.q0, flows=flows)
        self.nfm.to(device)
           
    def forward(self, x, train=True):
        log_det = self.q0.log_prob(x)
        z, log_det_forward = self.nfm.forward_and_log_det(x)

        min_values = torch.min(z, dim=0).values
        max_values = torch.max(z, dim=0).values
        normalized_data = (z - min_values) / (max_values - min_values)
        
        normalize_tensor = (self.hyper_range.to(self.device)).expand(z.size()[0], -1, -1) # output shape [task_batch, dim_z, 2]
        norm_z = normalize_tensor[:,:,0] * normalized_data + normalize_tensor[:,:,1] # normalize the transformed task into valid ranges
        
        log_det_norm = torch.sum(torch.log(normalize_tensor[:,:,0]/ (max_values - min_values)), dim=-1)

        z_reverse, loss = self.nfm.forward_kld(x)
            
        a, b = self.hyper_range[0][1], torch.sum(self.hyper_range[0])
        c, d = self.hyper_range[1][1], torch.sum(self.hyper_range[1])
        condition = (z_reverse[:, 0] >= a) & (z_reverse[:, 0] <= b) & (z_reverse[:, 1] >= c) & (z_reverse[:, 1] <= d)
            z_reverse = z_reverse[condition]
            
        if z_reverse.shape[0] == 0:
            log_det_reverse_total = torch.tensor(0.)
        else:
            log_det_z = self.q0.log_prob(z_reverse)
            log_det_reverse_total = -torch.mean(log_det_z) + loss
            
        return z, norm_z, log_det, log_det_forward, log_det_norm, log_det_reverse_total
        

# This part is to introduce the MLP for the implementation of MAML. --> Meta Player

class Meta_Learner(nn.Module):
    def __init__(self,
                 n_inputs,
                 n_outputs,
                 n_weights,
                 task_type,
                 device
                 ):
        '''
        n_inputs: the number of inputs to the network,
        n_outputs: the number of outputs of the network,
        n_weights: for each hidden layer the number of weights, e.g., [128,128,128]
        device: device to deploy, cpu or cuda
        '''
        
        super(Meta_Learner, self).__init__()

        # initialise lists for biases and fully connected layers
        self.weights = []
        self.biases = []

        # add one
        if task_type == 'sine':
            self.nodes_per_layer = n_weights + [n_outputs]
        elif task_type == 'acrobot':
            self.nodes_per_layer = n_weights + [n_outputs-2]
        elif task_type == 'pendulum':
            self.nodes_per_layer = n_weights + [n_outputs-1]

        # additional biases
        self.task_context = torch.zeros(0).to(device)
        self.task_context.requires_grad = True

        # set up the shared parts of the layers
        prev_n_weight = n_inputs
        for i in range(len(self.nodes_per_layer)):
            w = torch.Tensor(size=(prev_n_weight, self.nodes_per_layer[i])).to(device)
            w.requires_grad = True
            self.weights.append(w)
            b = torch.Tensor(size=[self.nodes_per_layer[i]]).to(device)
            b.requires_grad = True
            self.biases.append(b)
            prev_n_weight = self.nodes_per_layer[i]

        self._reset_parameters()

    def _reset_parameters(self):
        for i in range(len(self.nodes_per_layer)):
            stdv = 1. / math.sqrt(self.nodes_per_layer[i])
            self.weights[i].data.uniform_(-stdv, stdv)
            self.biases[i].data.uniform_(-stdv, stdv)

    def forward(self, x, task_type='sine'):
        x = torch.cat((x, self.task_context))

        for i in range(len(self.weights) - 1):
            x = F.relu(F.linear(x, self.weights[i].t(), self.biases[i]))
        
        if task_type == 'sine':
            y = F.linear(x, self.weights[-1].t(), self.biases[-1])
        elif task_type == 'acrobot':
            y = F.linear(x, self.weights[-1].t(), self.biases[-1])
            y = torch.cat((torch.cos(y[...,0:1]),torch.sin(y[...,0:1]),torch.cos(y[...,1:2]),torch.sin(y[...,1:2]),y[...,:2]),dim=-1)
        elif task_type == 'pendulum':
            y = F.linear(x, self.weights[-1].t(), self.biases[-1])
            y = torch.cat((torch.cos(y[...,0:1]),torch.sin(y[...,0:1]),y[...,:1]),dim=-1)            
        
        return y
\end{lstlisting}

As noted in the distribution adversary, the last layer is to normalize the range of the task identifiers into the pre-defined range, where the min-max normalization $\sigma(x)=\frac{x-x_{\text{min}}}{x_{\text{max}} - x_{\text{min}}}$ is utilized.
We consider the dimension of the task identifier to be $d$.
Let $a=\texttt{normalize\_tensor}[:,:,0]=[a_1,a_2,\dots,a_d]$ and $b=\texttt{normalize\_tensor}[:,:,1]=[b_1,b_2,\dots,b_d]$, the final layer of normalizing flows $g_m$ can be expressed as $\tau^{m}=g_m(\tau^{m-1})=a\odot\sigma(\tau^{m-1})+b$, where $\tau^{m}$ indicates the transformed task $\tau^m$ and the resulting sequence after $\{g_m\}_{m=1}^M$ is $\tau^0\mapsto\tau^1\mapsto\dots\mapsto\tau^M$.
Then, the resulting log-probability follows the computation:
\begin{equation}
\begin{split}
    \ln p(\tau^{m})=\ln p(\tau^{m-1})-\ln\Big|\det\frac{dg_{m}(\tau^{m-1})}{d\tau^{m-1}}\Big|\\
    =\ln p(\tau^{m-1})-\ln\Big(\prod_{i=1}^{d}\frac{a_i}{\tau^{{m-1,i}}_{\text{max}}-\tau^{{m-1,i}}_{\text{min}}}\Big)\\
    =\ln p(\tau^{m-1})-\Big(\sum_{i=1}^{d}\ln a_i-\ln (\tau^{{m-1,i}}_{\text{max}}-\tau^{{m-1,i}}_{\text{min}})\Big).
\end{split}
\end{equation}

\begin{equation}
    \begin{split}
        \mathbb{E}_{p_{\bm\phi_*}(\tau)}\Big[\mathcal{L}(\mathcal{D}_{\tau}^{Q},\mathcal{D}_{\tau}^{S};\bm\theta_*)\Big]
        \approx
        \frac{1}{K}\sum_{k=1}^{K}\mathcal{L}(D_{\tau_k^M}^{Q},D_{\tau_k^M}^{S};\bm\theta_*),
        \quad
        \text{with}
        \
        \tau_k^M=\texttt{NN}_{\bm\phi_*}(\tau_k^0)
        \
        \text{and}
        \
        \tau_k^0\sim p_{0}(\tau)
    \end{split}
\end{equation}

\subsection{Neural Architectures \& Optimization}

Our approach is meta-learning method agnostic, and the implementation is \textit{w.r.t.} MAML \citep{finn2017model} and CNP \citep{garnelo2018conditional} in this work.
As a result, we respectively describe the neural architectures in separate implementations and benchmarks.
We do not vary the neural architecture of meta learners, and only risk minimization principles are studied in experiments.

In the sinusoid regression and system identification benchmarks, we set the Lagrange multiplier $\lambda=0.2$ to cover most shifted task distributions.
All MAML-like methods employ a multilayer perceptron neural architecture. 
This architecture comprises three hidden layers, with each layer consisting of 128 hidden units. 
The activation function utilized in these models is the Rectified Linear Unit (ReLU).
The inner loop utilizes the stochastic gradient descent (SGD) algorithm to perform fast adaptation on each task, while the outer loop employs the meta-optimizer Adam to update the initial parameters of the model.
The learning rate for both the inner and outer loops is set to 1e-3.
As for CNP-like methods, the encoder uses a three-layer MLP with 128-dimensional hidden units and outputs a 128-dimensional representation. The output representations are averaged to form a single representation. This aggregated representation is concatenated with the query dataset and passed through a two-layer MLP decoder.
The optimizer used is Adam, with a learning rate of 1e-3.

In the continuous control benchmark, we set $\lambda=0.0$, and MAML serves as the underlying neural network architecture in our research.
The agent’s reward is the negative squared distance to the goal.
The agents are trained for one gradient update, employing policy gradient with the generalized advantage estimation \citep{Schulmanetal_ICLR2016} in the inner loop and trust-region policy optimization (TRPO) \citep{schulman2015trust} in the outer loop update.
The learning rate for the one-step gradient update is set to 0.1.

\subsection{Distribution Adversary Implementations}
The distribution adversary is implemented with the help of normalizing flows.
For AR-MAML, we adopt the distribution adversary with the neural network as follows.
We employ a 2-layer Planar flow \citep{rezende2015variational} to transform the initial distribution.
In each layer, the dimension of the latent variable is 2, and the activation function is leaky ReLU.
In the last layer, the min-max normalization is used. 
We use Adam with a cosine learning rate scheduler for the distribution adversary optimizer.

\section{Additional Experimental Results} \label{append: additional experimental results}

\begin{figure*}[h]
\begin{center}
\centerline{\includegraphics[width=0.9\textwidth]{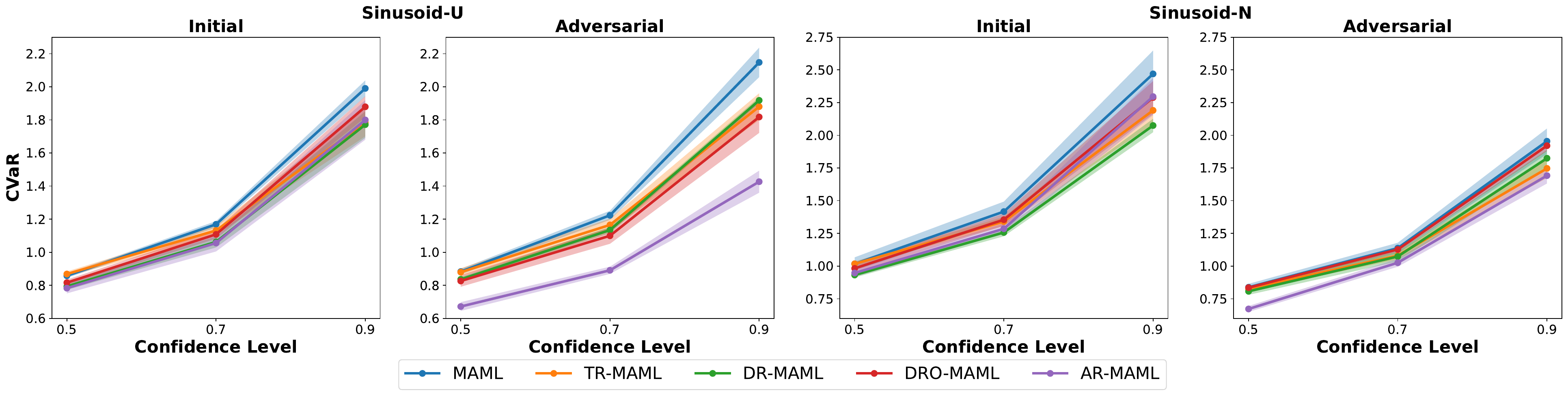}}
\vspace{15pt}
\caption{\textbf{$\text{CVaR}_{\alpha}$ MSEs with Various Confidence Level $\alpha$.}
Sinusoid-U/N denotes Uniform/Normal as the initial distribution type.
The plots report meta testing $\text{CVaR}_{\alpha}$ MSEs in initial and adversarial distributions with standard error bars in shadow regions.
}
\label{append: sinusoid_cvar}
\end{center}
\end{figure*}

\begin{figure*}[h]
\begin{center}
\centerline{\includegraphics[width=0.9\textwidth]{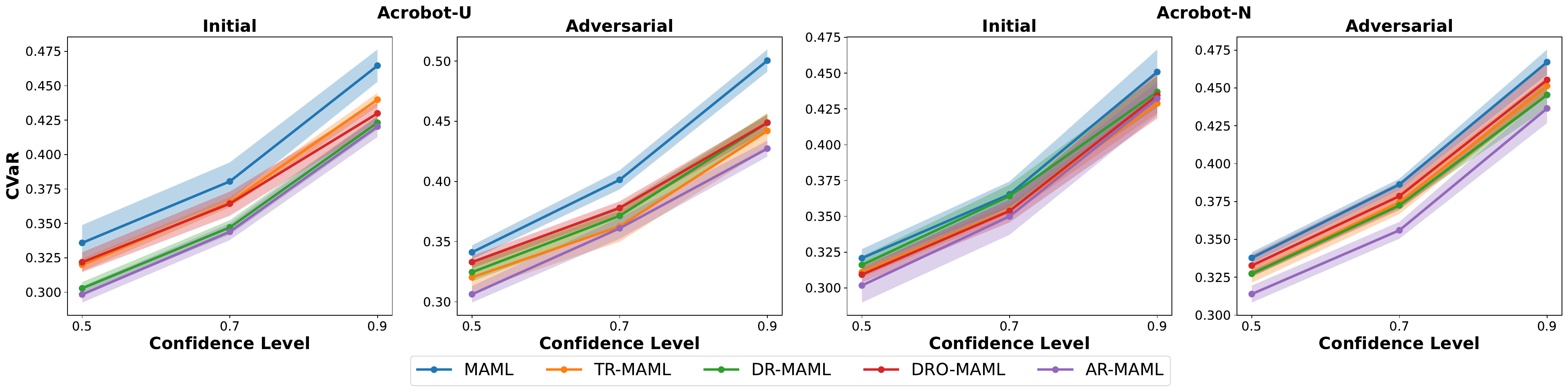}}
\vspace{15pt}
\caption{\textbf{$\text{CVaR}_{\alpha}$ MSEs with Various Confidence Level $\alpha$.}
Acrobot-U/N denotes Uniform/Normal as the initial distribution type.
The plots report meta testing $\text{CVaR}_{\alpha}$ MSEs in initial and adversarial distributions with standard error bars in shadow regions.
}
\label{append: acrobot_cvar}
\end{center}
\end{figure*}

\textbf{Tail Risk Robustness.}
Due to the page limit in the main paper, we present experimental results on tail risk robustness across various confidence levels $\alpha$ for sinusoid regression and acrobot system identification.
As illustrated in Figures \ref{append: sinusoid_cvar}/\ref{append: acrobot_cvar}, the advantage of AR-MAML is similar to that in pendulum system identification.
AR-MAML exhibits a more significant performance gain in the adversarial distribution than in the initial distribution.

\begin{figure*}[h]
\begin{center}
\centerline{\includegraphics[width=1.0\textwidth]{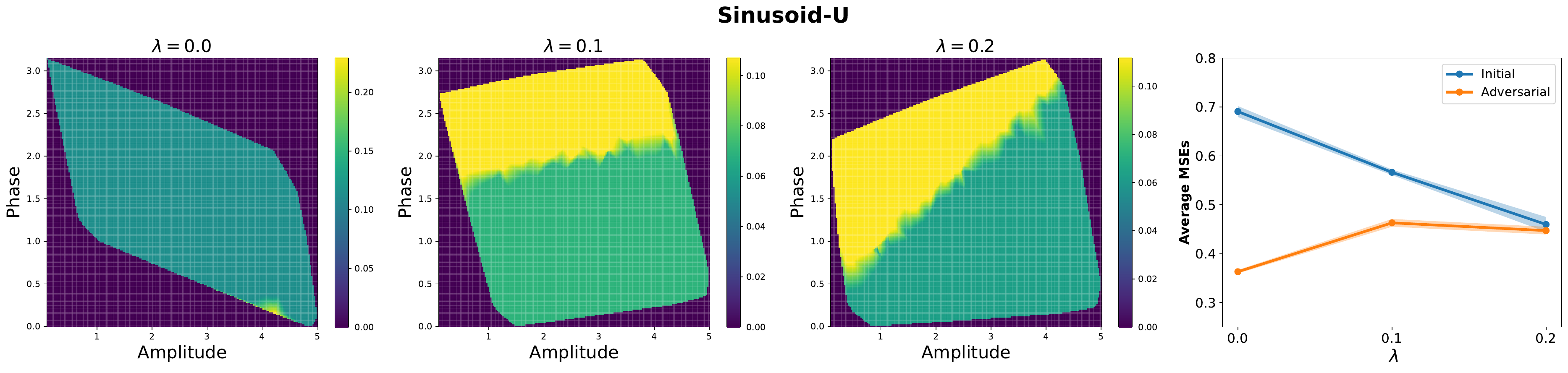}}
\vspace{15pt}
\caption{The first three plots show adversarial task probability distributions with varying Lagrange multipliers $\lambda$ in the sinusoid-U benchmark. The last plot depicts meta testing MSEs across different values of $\lambda$.
}
\label{append: lagrange_sinusoid_task_structure_performance}
\end{center}
\end{figure*}

\begin{figure*}[h]
\begin{center}
\centerline{\includegraphics[width=1.0\textwidth]{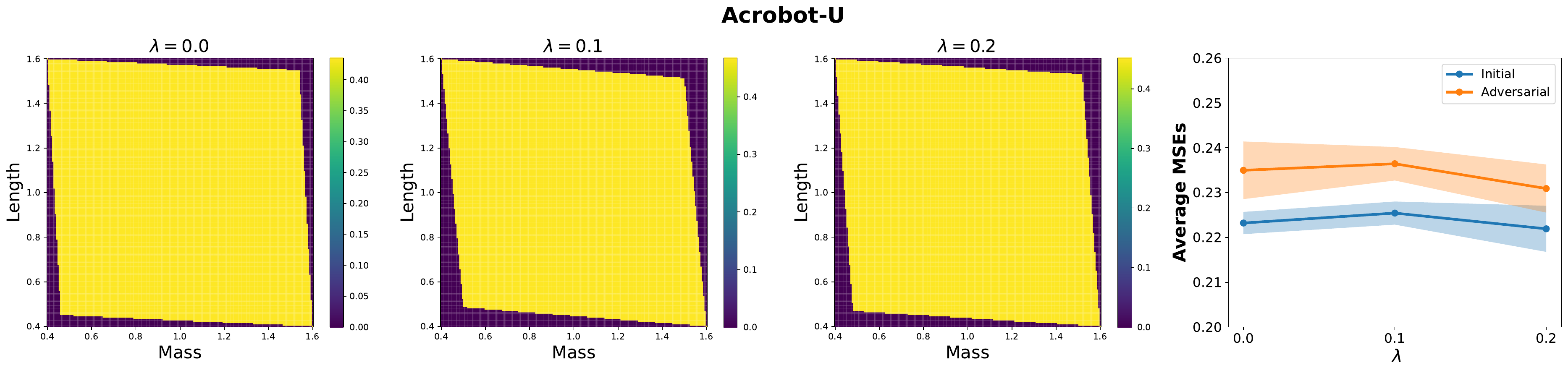}}
\vspace{15pt}
\caption{The first three plots show adversarial task probability distributions with varying Lagrange multipliers $\lambda$ in the acrobot-U benchmark. The last plot depicts meta testing MSEs across different values of $\lambda$.
}
\label{append: lagrange_acrobot_task_structure_performance}
\end{center}
\end{figure*}

\begin{figure*}[h]
\begin{center}
\centerline{\includegraphics[width=1.0\textwidth]{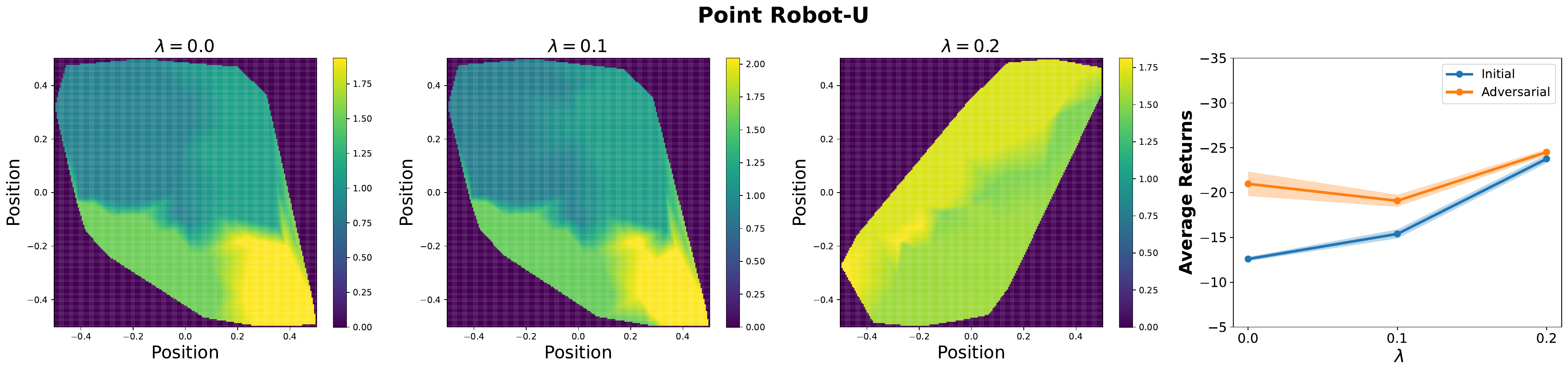}}
\vspace{15pt}
\caption{The first three plots show adversarial task probability distributions with varying Lagrange multipliers $\lambda$ in the point robot-U benchmark. The last plot depicts meta testing MSEs across different values of $\lambda$.
}
\label{append: lagrange_point_robot_task_structure_performance}
\end{center}
\end{figure*}

\textbf{Impacts of Shift Distribution Constraints.}
Still, by varying the Lagrange multipliers $\lambda$, we include the learned task structures and the meta-testing results in Figures \ref{append: lagrange_sinusoid_task_structure_performance}/\ref{append: lagrange_acrobot_task_structure_performance}/\ref{append: lagrange_point_robot_task_structure_performance}.
For sinusoid and acrobot cases, empirical findings are similar to those in the main paper.
In point robot navigation, we observe high probability density regions significantly change, and even the meta-testing results are improved in the initial task distribution with increasing $\lambda$ values. 

\setlength{\tabcolsep}{30.0pt}
\begin{table*}[ht]
\caption{\textbf{Entropy of initial distribution and adversarial distribution under different $\lambda$ values.} }
\centering
\begin{adjustbox}{max width = 1.0\linewidth}
\begin{tabular}{l|c|ccc}
\toprule
Benchmark & Meta-Test Distribution & $\lambda=0.0$ & $\lambda=0.1$ & $\lambda=0.2$\\
\bottomrule
\multirow{2}{*}{Sinusoid-U} & Initial & 2.734 & 2.734 & 2.734\\
& Adversarial & 2.15\scriptsize$\pm$0.01 & 2.44\scriptsize$\pm$0.01 & 2.46\scriptsize$\pm$0.01 \\
\multirow{2}{*}{Sinusoid-N} & Initial & 1.922 & 1.922 & 1.922\\
& Adversarial & -2.92\scriptsize$\pm$0.00 & 1.11\scriptsize$\pm$0.02 & 1.72\scriptsize$\pm$0.00 \\

\bottomrule
\end{tabular}
\end{adjustbox}
\label{sinusoid_entropy}
\end{table*}

\section{Platforms \& Computational Tools}

This project uses NVIDIA A100 GPUs in numeric computation. 
And we employ Pytorch \citep{paszke2019pytorch} as the deep learning toolkit in implementing experiments.

\end{document}